\documentclass[journal]{IEEEtran}

\usepackage{xcolor}
\usepackage{siunitx}
\usepackage{amssymb}
\usepackage{amsmath}
\usepackage{booktabs}
\usepackage[pdftex]{graphicx}
\usepackage{multirow}
 
\usepackage{xspace}
\usepackage{algorithmic}
\usepackage{algorithm}
\usepackage{hyperref}
\newcommand{\editorcomment}[1]{}

\begin{document}
\bstctlcite{BSTcontrol}

\title{Self-Adversarial Training incorporating Forgery Attention for Image Forgery Localization }

\author{Long~Zhuo, 
  Shunquan~Tan*,~\IEEEmembership{Senior Member,~IEEE,}
  Bin~Li,~\IEEEmembership{Senior Member,~IEEE,}
  and~Jiwu~Huang,~\IEEEmembership{Fellow,~IEEE}
  \thanks{*Corresponding author: Shunquan Tan.}  \thanks{All of the
    members are with the Guangdong Key Laboratory of Intelligent
    Information Processing, Shenzhen Key Laboratory of Media Security,
    Guangdong Laboratory of Artificial Intelligence and Digital
    Economy (SZ), Shenzhen Institute of Artificial Intelligence and
    Robotics for Society, China (email: zhuolong@email.szu.edu.cn;
    tansq, libin, jwhuang@szu.edu.cn). }
	
  \thanks{ S. Tan is with College of Computer Science and Software
    Engineering, Shenzhen University, Shenzhen 518060, China.}
  \thanks{This work was supported in part by NSFC (U19B2022, 61772349,
    61872244, 62072313, 61806131, 61802262), Guangdong Basic and
    Applied Basic Research Foundation (2019B151502001), and Shenzhen
    R\&D Program (JCYJ20200109105008228, 20200813110043002). This work
    was also supported in part by Alibaba Group through Alibaba
    Innovative Research (AIR) Program.}  }




\maketitle

\begin{abstract}
  Image editing techniques enable people to modify the content of an
  image without leaving visual traces and thus may cause serious
  security risks. Hence the detection and localization of these
  forgeries become quite necessary and challenging. Furthermore,
  unlike other tasks with extensive data, there is usually a lack of
  annotated forged images for training due to annotation
  difficulties. In this paper, we propose a self-adversarial training
  strategy and a reliable coarse-to-fine network that utilizes a
  self-attention mechanism to localize forged regions in forgery
  images. The self-attention module is based on a Channel-Wise High
  Pass Filter block (CW-HPF). CW-HPF leverages inter-channel
  relationships of features and extracts noise features by high pass
  filters. Based on the CW-HPF, a self-attention mechanism, called
  \textit{\textbf{forgery attention}}, is proposed to capture rich
  contextual dependencies of intrinsic inconsistency extracted from
  tampered regions. Specifically, we append two types of attention
  modules on top of CW-HPF respectively to model internal
  interdependencies in spatial dimension and external dependencies
  among channels. We exploit a coarse-to-fine network to enhance the
  noise inconsistency between original and tampered regions. More
  importantly, to address the issue of insufficient training data, we
  design a self-adversarial training strategy that expands training
  data dynamically to achieve more robust performance. Specifically,
  in each training iteration, we perform adversarial attacks against
  our network to generate adversarial examples and train our model on
  them. The proposed method is based on the assumption of content-changed manipulations. Extensive experimental results demonstrate that our proposed
  algorithm steadily outperforms state-of-the-art methods by a clear
  margin in different benchmark datasets.

\end{abstract}


%
\IEEEpeerreviewmaketitle

\section{Introduction}
\label{sec:intro}
%
%
%
%
\IEEEPARstart{T}{he} prevalence of digital image forgery is negatively affecting our lives, such as Internet rumors, insurance fraud, fake news, and even academic cheating~\cite{verdoliva_jstsp_2020}. Ghanim and Nabil~\cite{ghanim_icces_2018} revealed that image forgery might cause huge financial loss. Bik~\textit{et al.}~\cite{bik_mbio_2016} estimated that there were 3.8\% of 20,621 papers containing problematic figures with potential deliberate manipulation in biomedical research publications. Yet, the majority of image forgeries cases have not been detected~\cite{wu_cvpr_2019} due to the fact that image manipulation is very difficult to detect, and tampered images are commonly indistinguishable from original images by naked eyes. With the automatic generation techniques (e.g., inpainting with generative adversarial networks) and popular image editing tools (such as Photoshop$^\circledR$, After Effects Pro$^\circledR$, and the open-source software GIMP), it is easy to generate forged images at low cost. The localization of tampered regions turns out to be very tricky. Therefore, as tampered images grow at an enormous rate, it is necessary to develop new algorithms against image forgery.



Media forensics has been developed to detect image forgeries for many years. However, few early works focused on forgery detection at pixel level. The researchers made great efforts in classifying whether an image has been tampered or not with traditional features, like wavelet transform feature~\cite{muhammad_di_2012}, point matching~\cite{pun_tifs_2015}, and self-consistency~\cite{huh_eccv_2018}. Furthermore, most of them detect only one specific type of manipulation, such as splicing~\cite{huh_eccv_2018,salloum_jvcir_2018}, copy-move~\cite{wen_icip_2016}, and removal~\cite{zhu_spic_2018}.  Therefore, real-world media forensics is desperate for a new generation of algorithms that can obtain more refined results at the pixel level, as well as the detection of general manipulations.

However, it is challenging to localize multiple image forgeries at the pixel level since well-tampered images leave few visual traces. Conventional detection methods based on manually constructed statistical features rely heavily on the domain knowledge of human experts.

\begin{figure}[t]
	\centering
	\includegraphics[width=0.48\textwidth]{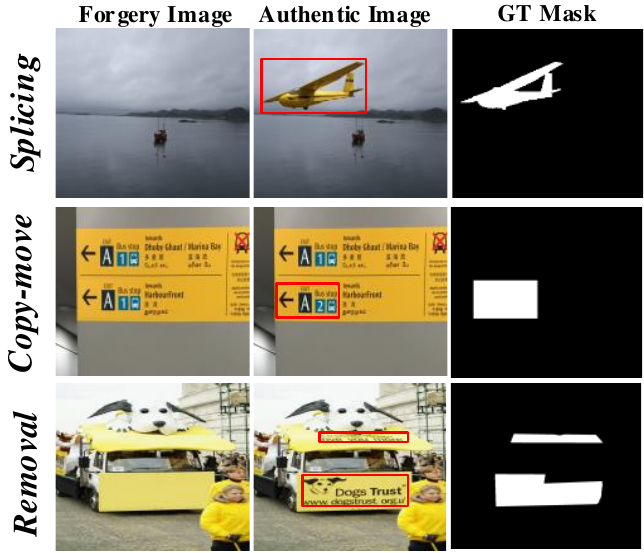}
	\caption{Examples of three common kinds of image forgeries, i.e., image splicing, copy-move and removal. Compared to authentic images, forgery images change the semantic meanings significantly. The tampered regions are highlighted with red rectangles in the authentic images. However, they are so secretive that it is hard for human eyes to point out them. }
	\label{type}
\end{figure}

Forgery localization task aims to localize the content-changed tampering techniques, including splicing, copy-move and removal, since content-changed techniques may cause serious misunderstandings while tampering with unchanged content does not change the semantics without causing misunderstandings. Please note that non-content-changed manipulations, such as
    Gaussian blur and JPEG compression, are excluded in the literature of
    image forgery localization, since those manipulations do not affect
    the semantic information expressed in the image scene. Therefore,
    all of the existing image forgery localization approaches~\cite{wu_cvpr_2019,zhou_cvpr_2018,bappy_cvpr_2017,bappy_tip_2019,hu_eccv_2020,zhuang_tifs_2021} only focus on three common semantic
    forgeries, i.e. image splicing, copy-move and removal. 
	Examples of the
    three common semantic forgeries are shown in Fig.~\ref{type}. Our work follows this assumption. 

Some early methods focused on low-level tampering artifacts in image forgeries, such as JPEG compression~\cite{bianchi_icassp_2011} and CFA color array analysis~\cite{ferrara_tifs_2012}.

Recent works, including Mantra-Net~\cite{wu_cvpr_2019}, RGB-N~\cite{zhou_cvpr_2018}, J-LSTM~\cite{bappy_cvpr_2017}, H-LSTM~\cite{bappy_tip_2019}, have proposed general and end-to-end solutions for image forgery localization using deep learning-based approaches. Both Wu~\textit{et al.}~\cite{wu_cvpr_2019} and Zhou~\textit{et al.}~\cite{zhou_cvpr_2018} borrowed three high pass filters from Steganalysis Rich Model~(SRM)~\cite{fridrich_tifs_2012} prior to their end-to-end framework to discover noise inconsistency between authentic and tampered regions. Specifically, Mantra-Net~\cite{wu_cvpr_2019} is a joint system that predicts manipulation traces for both image manipulation classification and forgery localization. It employs a VGG-based~\cite{simonyan_arxiv_2014} feature extractor to capture location information. Zhou~\cite{zhou_cvpr_2018} presented a two-stream network which consists of a regular RGB-based faster R-CNN stream and a parallel stream generated with SRM based noise extractor. The drawback of RGB-N is that it can only mark tampered regions with rectangular boxes due to the R-CNN structure. In the meantime, Bappy ~\textit{et al.}~\cite{bappy_cvpr_2017} applied an LSTM-based patch comparison method~(J-LSTM) to detect the boundary of tampered and authentic patches and further presented a separate encoder-decoder structure~(H-LSTM)~\cite{bappy_tip_2019} to improve the performance. 

More recently, some approaches~\cite{islam_cvpr_2020,zhu_tii_2020,hu_eccv_2020} used attention modules to focus on important regions in a target image. In the most recent one of them, Hu ~\textit{et al.}~\cite{hu_eccv_2020} proposed a so-called SPAN model (Spatial Pyramid Attention Network), which is also selected for comparison in our work. The problem of all of the mentioned attention-module based methods is that their key components are directly borrowed from object detection, and consequently are not well incorporated in forgery localization. For instance, their attention modules are object-sensitive and only pay attention to salient objects in an image. Since quite a few tampered regions are not objects in real scenarios, traditional object attention modules might fail to guide networks’ attention to tampered regions. Therefore, it is important to construct a novel network structure which can adapt attention mechanisms to forgery localization.

Furthermore, all state-of-the-art end-to-end solutions, e.g., SPAN~\cite{hu_eccv_2020}, Mantra-Net~\cite{wu_cvpr_2019}, and RGB-N~\cite{zhou_cvpr_2018}, have adopted normal data augmentation techniques, such as image flipping and rotation, to obtain twice or more number of training samples to avoid model over-fitting. However, with ultra-limited training samples in this field, additional samples obtained with normal data augmentation are still limited. It is necessary to develop a new training strategy for forgery localization to increase sample numbers by one or two orders of magnitude.

To address the above issues, we propose a novel framework and a self-adversarial training strategy toward the goal of localizing forged regions in images precisely. The proposed network involves a Channel-Wise High Pass Filter (CW-HPF) block and a \textit{\textbf{forgery attention}} mechanism. Furthermore, we introduce the coarse-to-fine architecture to enhance network representations. 

Specifically, CW-HPF splits an input feature map into channels. It then performs convolution initialized by three high pass filters, called HPF-Conv, on each channel to extract noise features, respectively. The HPF-Conv output is fused to uncover the inter-channel relationships of these noise features. The channel-wise architecture of CW-HPF is motivated by WISERNet~\cite{zeng_tifs_2019}, a powerful deep steganalysis model using inter-channel information.

On top of CW-HPF, we propose \textit{\textbf{forgery attention}}, a
novel attention mechanism considering forgery localization
task. Inspired by the dual attention module~\cite{fu_cvpr_2019},
\textit{\textbf{forgery attention}} aims to capture noise feature
dependencies in both spatial and channel dimensions. In the spatial
dimension, we extract spatial dependencies of noise features in every
pixel pair of feature maps. Similarly, in the channel dimension, we
capture channel dependencies of noise features among channel
pairs. The outputs of two attentive dimensions are fused to enhance
feature representations. \textit{\textbf{Forgery attention}} enables
our model to adaptively integrate local noise features with their
global dependencies.

To alleviate the problem of limited training data, we present a novel Self-Adversarial Training~(SAT) strategy for dynamical data augmentation. SAT exploits adversarial attacks~\cite{goodfellow_arxiv_2014} in every training iteration and generates new training data dynamically, which guides our model to defend from adversarial attacks and achieve more robust performance. Different from traditional data augmentation, our SAT can provide inexhaustible new adversarial training data according to model updating.

As for the experiments, the notable public datasets for forgery localization, including DEFACTO~\cite{mahfoudi_espc_2019}, NIST~\cite{guan_wacvw_2019}, Columbia~\cite{mahfoudi_espc_2019}, COVERAGE~\cite{dong_csicsip_2013}, CASIA~\cite{wen_icip_2016} and PS-dataset~\cite{zhuang_tifs_2021}, are established on the content-changed manipulations. Therefore, the assumption of our experiments is that the forgery localization models aim to localize the content-changed regions. Based on this assumption, extensive experiments demonstrate that our approach has achieved state-of-the-art results on all primary benchmark datasets.

To be summarized, our main contribution is to propose a novel image	deep-learning based forgery localization framework that organically combines the domain knowledge of multimedia security and image recognition. The novelty of this work can be explicitly highlighted as follows: 1) we propose \textbf{\textit{Forgery attention}}, a novel attention mechanism for forgery localization, that fuses domain knowledge well-established in
	multimedia security into attention mechanism, and makes a
	combination of channel-wise and spatial dependencies.
	2) A new training strategy, namely SAT, is first introduced to alleviates the problem of the lack of training data for improving the localization performance and the model robustness.
	3) we move a further step for HPF layer and present CW-HPF that exploits the inter-channel relationships and generates more precise noise features.

The remainder of the paper is organized as follows. In Section~\ref{sec:methodology}, we present our proposed approach in details. Then, we show the results of our extensive experiments in Section~\ref{sec:experiment}. Finally, we conclude this paper in Section~\ref{sec:conclu}.

\begin{figure*}[t]
	\centering
	\includegraphics[width=0.99\textwidth]{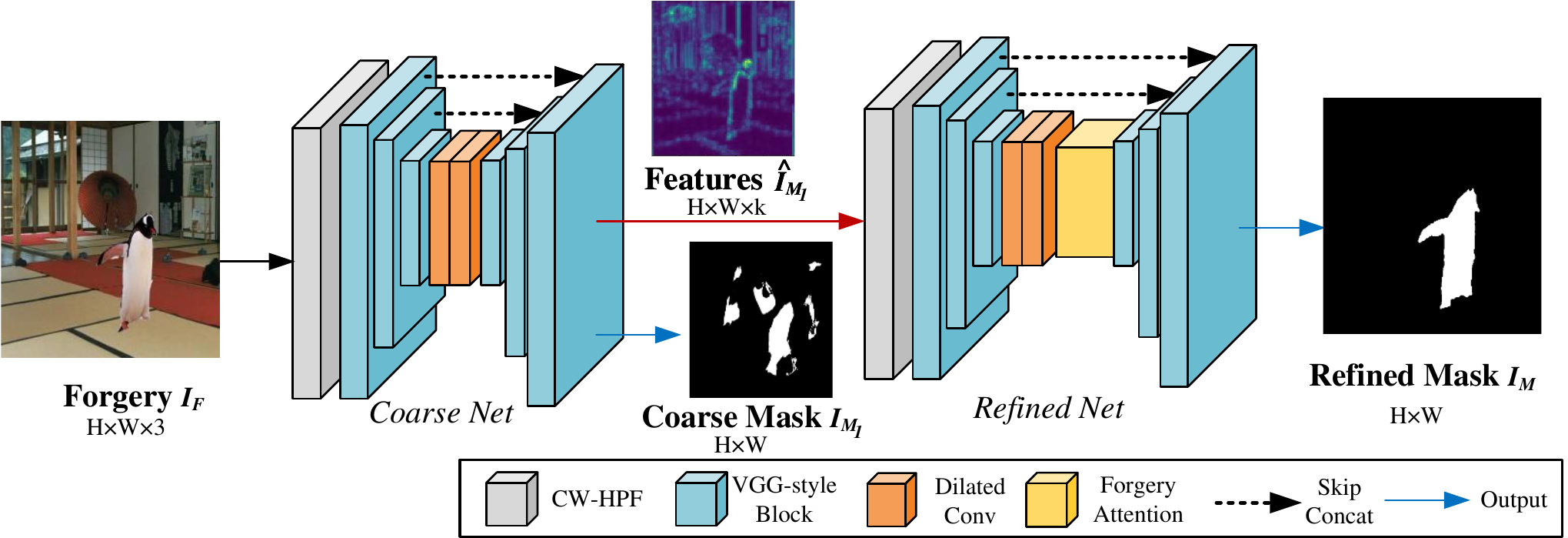}
	\caption{Illustration of our proposed coarse-to-fine network for forgery localization task. It is composed of two sub-networks, namely coarse-net and refined-net. The coarse-net takes a forged image~$I_{F}$ as input and predicts a coarse mask~$I_{M_1}$ and a feature map~$\hat{I_{M_1}}$. The refined-net takes~$\hat{I_{M_1}}$ and predicts a refined mask~$I_M$.}
	\label{net}
\end{figure*}

\section{Proposed Approach}
\label{sec:methodology}
In this section, we present the general framework of our proposed network as well as self-adversarial training strategy and then formulate our approach. 

\subsection{Overview}
Given an image, we aim to localize its tampered regions at the pixel level. The general framework of our proposed model is illustrated in Fig.~\ref{net}. Furthermore, we propose self-adversarial training to promote the robustness of the network since forged training samples are ultra-limited, whose details will be given in Sect.~\ref{sec:sat}. Overall, we construct a coarse-to-fine network containing a coarse output mask and a refined output mask. Next, we propose a two-phase training strategy called self-adversarial training strategy to perform our training process, where our network is trained using the original input-output pairs commonly in the first phase and then trained using new input-output pairs with adversarial samples in the second phase. 

Our network has a coarse-to-fine architecture. It is composed of two sub-networks, i.e., a \textbf{\textit{coarse net}} and a \textbf{\textit{refined net}}. The \textit{coarse net} is fed with a forged image~$I_F$ of size~$H\times W \times 3$ and outputs a coarse prediction mask of size~$H\times W $ and a feature map~$\hat{I_{M_1}}$ of size~$H\times W\times k$, where~$k$ is set to 16 because the output of the last block outputs a feature map with 16 channels. The feature map~$\hat{I_{M_1}}$ is fed into the \textit{refined net} that predicts a refined prediction mask~$I_M$ of size~$H\times W$. Note that~$I_{M_1}$ and~$I_M$ come from two output convolution layers with a kernel size of~$7\times 7$, activated by a sigmoid function. $\hat{I_{M_1}}$ aims to deliver complete feature information to the \textit{refined net}, which enables the \textit{refined net} to be optimized along with the features. The predicted masks mark tampered regions in white and leave the rest in black. We select the refined mask~$I_M$ as the final result.

As illustrated in Fig.~\ref{net}, there are three key components in the \textit{coarse net}, including a CW-HPF block, multiple VGG-style blocks, and a series of dilated convolutional layers. The \textit{refined net} adopts similar structure, but with a \textit{\textbf{forgery attention}} module attached.

\subsection{CW-HPF Block}

HPF layer widely used in image steganalysis~\cite{fridrich_tifs_2012} has been adopted in
forgery localization~\cite{wu_cvpr_2019,zhou_cvpr_2018}. The intuition
behind the usage of high-pass filters is that tampering traces are
generally manifested in the middle and high-frequency sub-bands of the
tampered image. However, prior works have neglected the relationships
among channels of images, which may yield many unnecessary noise
features. To extract more accurate noise features, we leverage the
inter-channel information to enhance the noise inconsistency between
tampered and pristine regions.WISERNet~\cite{zeng_tifs_2019}, our prior work in the field
of image steganalysis has proved in theory that noise features
extracted by channel-wise high pass filters could enlarge the slight
difference between the authentic images and the manipulated images
with tiny stego noises. However, WISERNet only uses the channel-wise
high pass filters with fixed weights and three R-G-B color channels
as input in a bottom pre-processing layer.

Move a step further, in our proposed framework we have introduced
CW-HPF which can be fed with arbitrary
input channels and can be put in arbitrary positions/branches of our
framework. 

\begin{figure}
	\centering
	\includegraphics[width=0.49\textwidth]{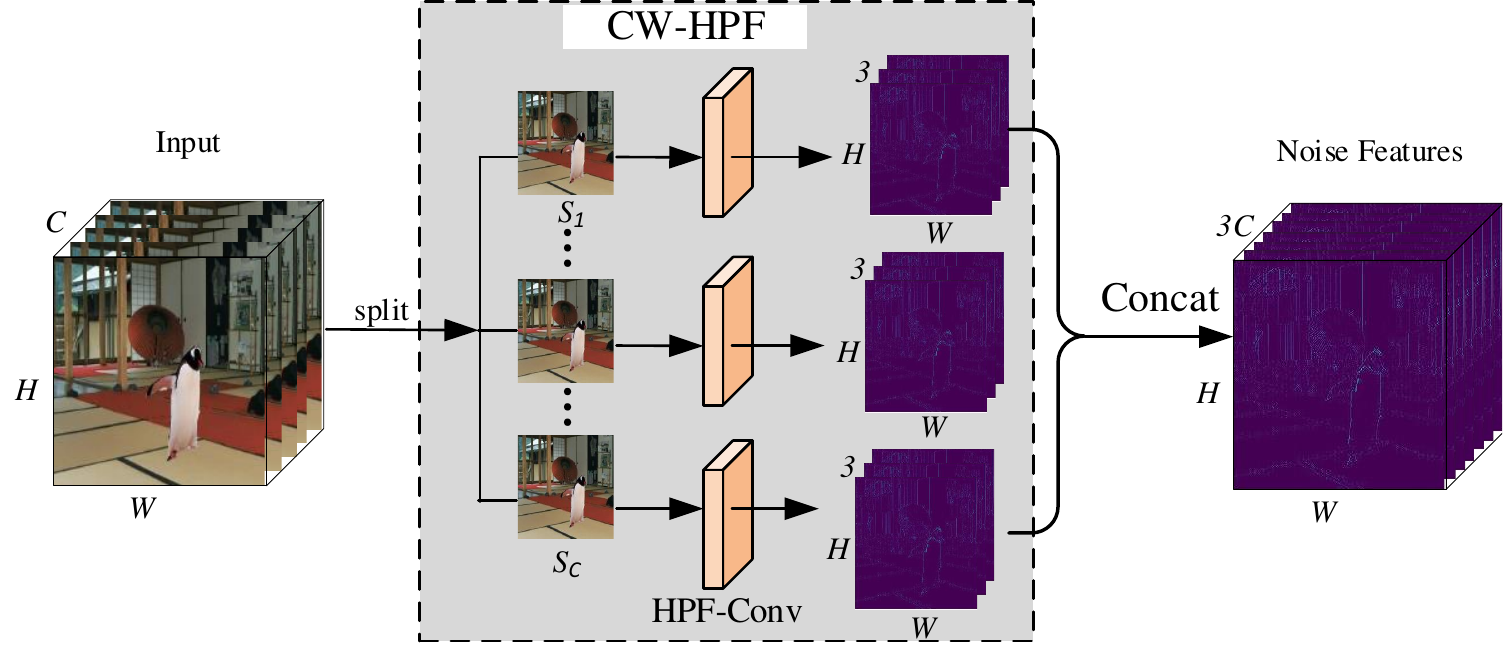}
	\caption{Details of the Channel-Wise High Pass Filter (CW-HPF) block. }
	\label{cw-hpf}
\end{figure}

\begin{figure}
	\centering
	\includegraphics[width=0.49\textwidth]{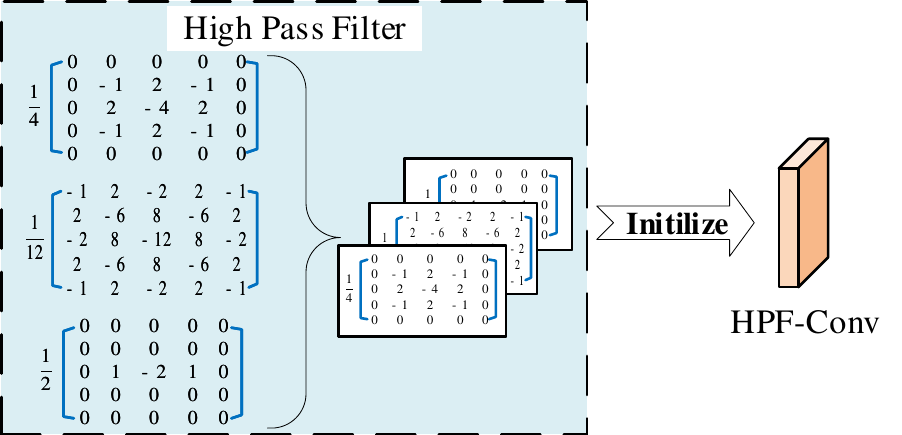}
	\caption{Details of the High Pass Filter Convolution (HPF-Conv) layer.}
	\label{filter}
\end{figure}

Fig.~\ref{cw-hpf} shows the details of the CW-HPF block. CW-HPF takes a feature map of size~$H\times W\times C$ and outputs a noise feature map of size~$H\times W\times 3C$. Firstly, the input feature map is converted to a feature map set~$\mathcal{S}$ with $C$ feature maps. $\mathcal{S}_i$ represents the~$i$-th channel of the input feature map. We define~$\mathcal{S}$ as~$\mathcal{S}=\{\mathcal{S}_1, \mathcal{S}_2, \mathcal{S}_3, \dots, \mathcal{S}_C\}$. We employ three high pass filters originated from SRM~\cite{fridrich_tifs_2012} to initialize a convolution layer of size~$5\times 5\times 3$, which we call HPF-Conv. As shown in Fig.~\ref{filter}, the selected high pass filters involve a KB filter, a KV filter and a first order kernel, which are the same as RGB-N~\cite{zhou_cvpr_2018}. Unlike steganalysis, the forgery localization task needs only three high pass filters from thirty steganalysis rich model filters to achieve decent performance and save computing resources~\cite{zhou_cvpr_2018}. We apply HPF-Conv to perform convolution with each element of the feature map set~$\mathcal{S}$, and the results are concatenated to produce noise features of size~$H\times W\times 3C$. 

Our motivation is that the slight high-frequency differences
	between the authentic images and the manipulated images ought to be
	passed on through cascaded layers. As a result, CW-HPF can
	be used in different parts of our proposed framework to extract more
	precise noise features by leveraging the inter-channel relationships,
	and thus improves the overall performance.

CW-HPF in this work is unlearnable, and the filters are
handcrafted.

\begin{figure*}
	\centering
	\includegraphics[width=1.0\textwidth]{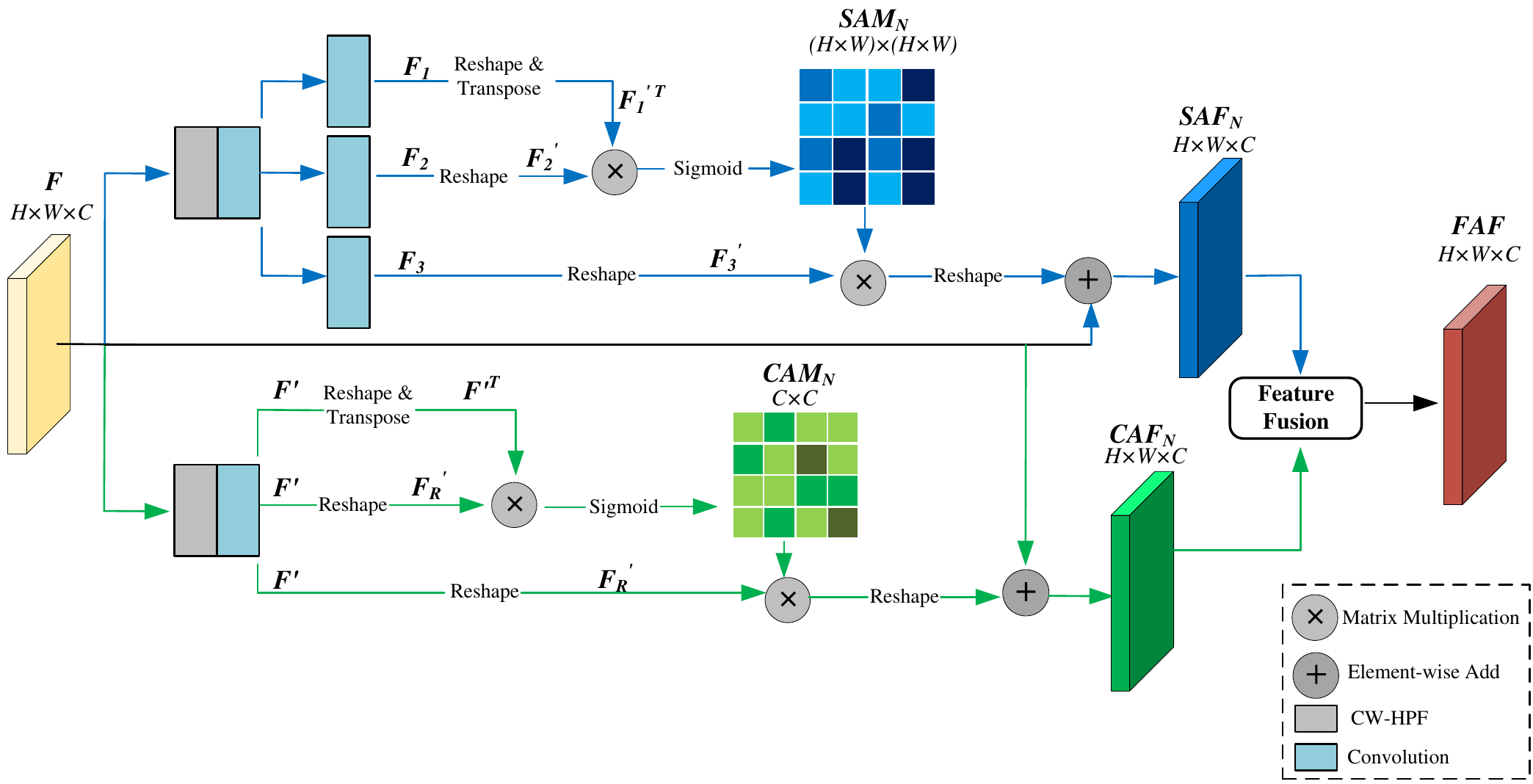}
	\caption{The details of \textit{\textbf{forgery attention}}. There are two parallel branches in \textit{\textbf{forgery attention}}. The top one is the \textit{spatial attention branch}, and the bottom one is the \textit{channel attention branch}. Note that the cells with different darkness in~$\text{SAM}_N$ and~$\text{CAM}_N$ indicate the amount of attention paid to the noise features of these regions. }
	\label{fa}
\end{figure*}
\subsection{Forgery Attention}

Attention mechanism enables a neural network focus on important regions of its feature representations. Attention helps it build input-aware connections to focus more on meaningful regions by replacing fixed weights with input dependent weights. Attention mechanism has been heavily involved in different application fields of deep-learning frameworks, such as machine translation~\cite{vaswani_nips_2017}, image captioning~\cite{xu_icml_2015} and object detection~\cite{li_cvpr_2019}. 

 SPAN~\cite{hu_eccv_2020} borrowed a single spatial attention mechanism
from image recognition, with two downsides:
1) it ignores the color-channel-wise dependencies, which are
	important for forgery localization of true color images with
	multiple color channels;
2) it only focuses on salient objects on the scene while quite a
	few tampered regions are not with salient objects.

To overcome the downsides of normal attention mechanism in forgery
localization, we have fused domain knowledge well established in
multimedia security into attention mechanism, as well as made a
combination of channel-wise and spatial dependencies. Here, we propose a \textit{\textbf{forgery attention}} mechanism to focus on tampered traces instead of salient objects in order to make it adapt to forgery localization tasks. The details of our forgery attention are formulated in Algorithm~\ref{alg1}.

We construct two attention branches to obtain global noise attention features, as shown in Fig.~\ref{fa}. We feed local features~$F \in \mathbb{R}^{H\times W\times C}$ generated by the dilated convolutional module of the \textit{refined net} into two CW-HPF-based parallel attention branches, namely a \textbf{\textit{spatial attention branch}} and a \textbf{\textit{channel attention branch}}.

Our motivation is that with our proposed forgery attention
module, the extracted attention features can provide the similarity
map of noise features in both channel and spatial dimensions. It can
reflect long-term contextual information in the noise domain since any
two positions with similar noise features can contribute mutual
improvement regardless of their distance in both spatial dimension and
channel dimension.

The \textit{spatial attention branch} generates a Spatial-dimension Attention Feature map~( denoted as~$\text{SAF}_N$). It draws the spatial relationship between pairwise positions of the noise features. We update the features of each position by aggregating noise features of all positions with a weighted sum, where the weights are calculated by the similarities of the noise features between the corresponding two positions. For any two positions with similar noise features in the spatial dimension, they can contribute to mutual improvement.

\begin{algorithm}[t]
	\renewcommand{\algorithmicrequire}{\textbf{Input:}}
	\renewcommand{\algorithmicensure}{\textbf{Output:}}
	\footnotesize
	\caption{\textbf{Forgery Attention Mechanism}}
	\label{alg1}
	\begin{algorithmic}[1]
		\REQUIRE the feature map $F\in \mathbb{R}^{H\times W\times C}$;
		\ENSURE the Forgery Attention Feature $FAF\in \mathbb{R}^{H\times W\times C}$;
		\STATE Initialize learnable parameters $\delta$ and $\gamma$;
		\FOR {every training iteration}
		\STATE \textit{Spatial attention branch}: \\
		\begin{enumerate}
			\item Generate noise features using CW-HPF and a followed convolution layer;
			\item Generate $F_1,F_2,F_3$ using three parallel convolution layers;
			\item Reshape $F_1$ into $\mathbb{R}^{(H\times W)\times C}$;
			\item Transpose the result of 3) $\rightarrow F'^\top_1 \in \mathbb{R}^{C\times (H\times W)}$;
			\item Reshape $F_2 \rightarrow F'_2 \in \mathbb{R}^{(H\times W)\times C}$;
			\item $Sigmoid(F'_2\times F'^\top_1)\rightarrow SAM_N \in \mathbb{R}^{(H\times W)\times (H\times W)}$;
			\item Reshape $F_3\rightarrow F'_3 \in \mathbb{R}^{(H\times W)\times C}$;
			\item Reshape $\delta \cdot (SAM_N\times F'_3)$ into $\mathbb{R}^{H\times W\times C}$;
			\item Element-wise addition of the result of 8) and $F\rightarrow SAF_N \in \mathbb{R}^{H\times W\times C}$;
		\end{enumerate}
		\STATE \textit{Channel attention branch}: \\
		\begin{enumerate}
			\item Generate noise features~$F'$ using CW-HPF and a followed convolution layer;
			\item \begin{enumerate}
				\item Reshape $F' \rightarrow F'_R \in \mathbb{R}^{(H\times W)\times C}$;
				\item Transpose $F'_R \rightarrow F'^\top \in \mathbb{R}^{C\times (H\times W)}$;
				
				\item $ Sigmoid(F'^\top \times F'_R)\rightarrow CAM_N\in \mathbb{R}^{C\times C}$;
				\item Reshape $\gamma \cdot (CAM_N \times F'_R)$ into $\mathbb{R}^{H\times W\times C}$;
				\item Add the result of d) and $ F \rightarrow CAF_N\in \mathbb{R}^{H\times W\times C}$;
			\end{enumerate}
		\end{enumerate}		
		\STATE $SAF_N+CAF_N \rightarrow FAF$;
		\STATE Update $\delta$ and $\gamma$ with back-propagation.
		\ENDFOR
		\RETURN $FAF$ 	
	\end{algorithmic}
\end{algorithm}

Specifically, in the spatial dimension, the input~$F$ is fed into the CW-HPF block, followed by a convolution layer, to extract noise features. Three convolution layers are applied over the noise feature map in parallel to generate three feature maps~${F_1,F_2,F_3}\in \mathbb{R}^{H\times W\times C}$, respectively. $F_1,\ F_2$ and~$F_3$ are then reshaped to two-dimensional feature maps~$F'_1,\ F'_2$ and $F'_3$, each of which belongs to $\mathbb{R}^{(H\times W)\times C}$. $F'_1$ is further transposed to $F'^\top_1 \in \mathbb{R}^{C\times(H\times W)}$. Matrix cross product of~$F'_2$ and~$F'^\top_1$ is performed to calculate the distances between different positions. Here, the result of the product measures the impact of $i^{th}$ position on $j^{th}$ position, where $(i,j)$ in $((H\times W), (H\times W))$. The matrix cross product is further fed to a Sigmoid activation to get the Spatial Attention Matrix of noise features~$\text{SAM}_N \in \mathbb{R}^{(H\times W)\times (H\times W)}$. $\text{SAM}_N$ describes the similarity between any two different positions in the spatial dimension of the noise feature map. According to~\cite{fu_cvpr_2019}, the more similarity between two positions in
the noise feature map is, the greater the correlation between
two points in $F$ becomes. $\text{SAM}_N$ is then multiplied with $F_3$, and the result is reshaped back to~$H\times W\times C$. Finally, a learnable scaling factor~$\gamma$ is multiplied with the result of the last step and then is added with the input features~$F$ to generate~$\text{SAF}_N \in \mathbb{R}^{H\times W\times C}$. It is formulated as follows:
\begin{equation}
	\label{saf}
	\begin{split}
		\text{SAF}_N&= \gamma(\text{SAM}_N\times F_3)+F \\
		&= \gamma(\text{sigmoid}(F^\top_1\times F'_2)\times F'_3)+F,
	\end{split}
\end{equation}
where we initialize~$\gamma$ as 0 and update it with back-propagation learning.

Meanwhile, the \textit{channel attention branch} generates a Channel-dimension Attention Feature map~( denoted as~$\text{CAF}_N$) to model the channel relationship between any two channels of noise features. We update each channel feature map with a weighted sum of all channel feature maps. 

The calculation in the channel dimension is similar to that in the spatial dimension. Firstly, a CW-HPF and a convolution layer are used to generate the noise features~$F'$ of input~$F$, where $F' \in \mathbb{R}^{H\times W\times C} $. $F'$ is reshaped to~$F_R'\in \mathbb{R}^{(H\times W)\times C}$. $F_R'$ is then performed matrix cross product with~$F'^\top$. A Sigmoid activation is applied to calculate the Channel Attention Matrix~$\text{CAM}_N \in \mathbb{R}^{C\times C}$ of noise features. Like~$\text{SAM}_N$, $\text{CAM}_N$ describes the similarity between any two different positions in the channel dimension of the noise feature map. Then~$F_R'$ is performed matrix cross product with~$\text{CAM}_N$, whose result is reshaped to~$H\times W\times C$. Finally, the matrix cross product result is multiplied by another learnable scaling factor~$\delta$, and the multiplication result is added with the input~$F$ to obtain~$\text{CAF}_N \in \mathbb{R}^{H\times W\times C}$:
\begin{equation}
	\label{caf}
	\begin{split}
		\text{CAF}_N&= \delta(\text{CAM}_N\times F_R')+F, \\
		&= \delta(\text{sigmoid}(F'^\top\times F_R')\times F_R)+F,
	\end{split}
\end{equation}
where~$\delta$ starts from 0 and gradually adjusts to assign more weight during training.

The introduction of the two learnable scaling factors, namely~$\gamma$ and $\delta$, during the training procedure of $\text{SAF}_N$ and $\text{CAF}_N$ is borrowed from~\cite{fu_cvpr_2019}, in order to enhance network representation. In addition, please note that there are only two classes (i.e., forged or not) in the forgery localization task, and usually prediction of every pixel ranges from 0 to 1. Therefore, in~\textit{\textbf{forgery attention}}, $\text{SAF}_N$ and~$\text{CAF}_N$ are generated with Sigmoid activation rather than traditional Softmax function since the value of Sigmoid activation is confined in [0,1].

After that, $\text{SAF}_N$ and~$\text{CAF}_N$ are then fused to obtain the Forgery Attention Features (FAF). In particular, an element-wise addition is performed on~$\text{SAF}_N$ and~$\text{CAF}_N$, of which the result is fed into a convolution layer to generate~$\text{FAF}$. As a result, FAF provides the similarity map of noise features in both channel and spatial dimensions. It reflects long-term contextual information in the noise domain because any two positions with similar noise features can contribute mutual improvement regardless of their distance in both spatial dimension and channel dimension. 


\subsection{VGG Block and Dilated Convolutional Block}
The architecture of VGG blocks originates from Mantra-Net~\cite{wu_cvpr_2019}. Each VGG block contains three or four stacked convolution layers with kernel size~$3\times 3$. In \textit{coarse net}, the VGG modules are denoted as $\mathcal{V}_i(f)\ ( i\in \{ 1,2,3,4,5\})$, where~$f$ denotes the input features and $\mathcal{V}_i$ denotes VGG block. $\mathcal{V}_i$ performs encoding when $i= 1, 2, 3$ while performs decoding when $i=4,5$. The VGG blocks are of size~$32 \times 2^i$  when $i= 1, 2, 3$, and size~$32 \times 2^{5-i}$ when $i=4,5$. Max pooling layers follow~$\mathcal{V}_1(f)$ and~$\mathcal{V}_2(f)$ to down-sample the features, and the output of them is skip-concatenated with that of~$\mathcal{V}_5(f)$ and~$\mathcal{V}_4(f)$, respectively. $\mathcal{V}_4(f)$ and~$\mathcal{V}_5(f)$ are followed by up-sampling layers to restore the size of feature maps.

Zhuang \textit{et al.}~\cite{zhuang_tifs_2021} has pointed out that for forgery localization task the deep-learning framework needs larger receptive fields to avoid learning features from narrow local regions. Thus following their approach, four dilated convolution layers are applied to inflate the kernels by inserting zeros between kernel elements with different dilation rates for extracting features with larger receptive fields. The dilated convolution layers are then used to bridge the encoders and the corresponding decoders in both the \textit{coarse net} and the \textit{refined net}. Specifically, the dilation rates in the four dilated convolution layers are 2, 4, 8, and 16, respectively.

\begin{figure*}
	\centering
	\includegraphics[width=1\textwidth]{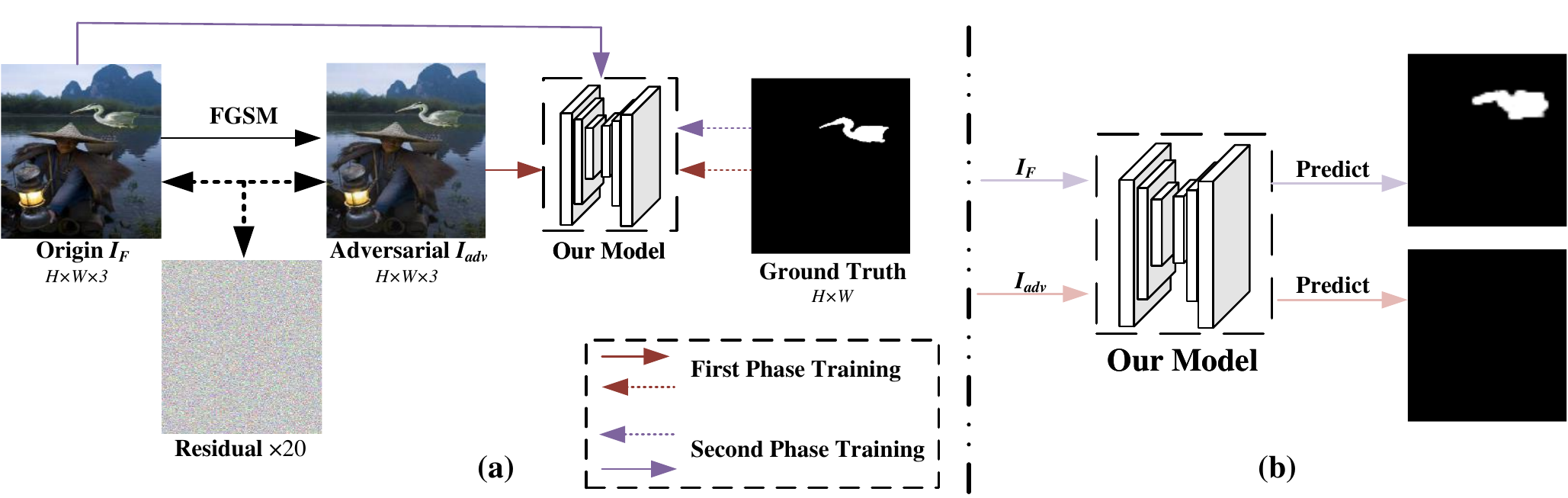}
	\caption{Illustration of our proposed SAT. It trains our model
          in two phases. Fig. (a) shows the training procedure of
          SAT. Values of the residual map between $I_F$ and $I_{adv}$
          are multiplied by 20 for better visualization. Fig. (b)
          shows different responses of our model prior to the second phase, when it is fed with $I_F$ or fed with $I_{adv}$. The network predicts a totally black mask input $I_{adv}$ while it predicts a roughly accurate mask input $I_F$. }
	\label{sat}
\end{figure*}

\begin{algorithm}[!t]
	\renewcommand{\algorithmicrequire}{\textbf{Input:}}
	\renewcommand{\algorithmicensure}{\textbf{Output:}}
	\footnotesize
	\caption{\textbf{Self-adversarial Training Strategy}}
	\label{alg2}
	\begin{algorithmic}[1]
		\REQUIRE The tampered image~$I_F$ and~$y_{gt}$, the corresponding binary mask of ground truth;
		\ENSURE The parameters~$\theta$ of our network;
		\STATE Initialize the parameters~$\theta$, including weights and biases, of our network as illustrated in Fig.~\ref{net};
		
		\FOR {every training iteration}
		\STATE \textit{\textbf{(Begin Phase 1 )}}
		\STATE Predict the coarse prediction mask~$I_{M_1}$ and the refined prediction mask~$I_M$ for the forged image~$I_F$;
		\STATE  $[I_F,(I_{M_1}, I_M),(y_{gt},y_{gt})]$ $\Rightarrow$FirstTrainSet
		\STATE Update parameters~$\theta$ of our network with FirstTrainSet;
		\STATE \textit{\textbf{(End Phase 1 )}}
		\STATE \textit{\textbf{(Begin Phase 2)}}
		\STATE $Random (0,0.01] \rightarrow  \epsilon$;
		\STATE Initialize FGSM algorithm~(Equation~\ref{eq:fgsm}) with $\epsilon$ and the updated $\theta$;
		\STATE $ FGSM(I_F,y_{gt})\rightarrow I_{adv}$
		
		\STATE Predict the coarse prediction mask~$I_{M_1}$ and the refined prediction mask~$I_M$ for the adversarial image~$I_{adv}$;
		\STATE  $[I_{adv},(I_{M_1},I_M),(y_{gt},y_{gt})]$$\Rightarrow$SecondTrainSet 
		\STATE Update parameters~$\theta$ of our network with SecondTrainSet;
		\STATE \textit{\textbf{(End Phase 2 )}}
		\ENDFOR
		
	\end{algorithmic}
\end{algorithm}

\subsection{Self-Adversarial Training}
\label{sec:sat}

Frankly speaking, deep-learning based models are all training data
hungry. It is a huge challenge to train a deep-learning based
framework in a scenario with limited training samples, such as image
forgery localization. Data augmentation techniques can alleviate this
issue. However, the common data augmentation techniques
, such as image flipping and rotation, used in
existing forgery localization methods~(e.g.
Mantra-Net~\cite{wu_cvpr_2019} and SPAN~\cite{hu_eccv_2020}) do not
utilize the feedback of the target deep-learning based framework in
training samples augmentation.

Our proposed SAT strategy can realize training samples augmentation
with endless supply of adversarial samples generated with the latest
gradients of the target deep-learning model in every training
iteration.

The motivation behind SAT is that the detection model can easily over-adapt to the texture features of the image datasets it is trained on, since the tampering noise is very subtle. SAT not only increases the robustness of the model with adversarial attack during training, but also improves its performance by providing training data dynamically. In the field of object detection, YOLO v4~\cite{bochkovskiy_arxiv_2020} has used self-adversarial training to augment the training data and achieve better performance. We first attempt for applying self-adversarial training strategy and making experimental analysis of it for forgery localization.

In the first training phase, like the traditional training process, our model is trained on the forged image~$I_F \in\mathbb{R}^{H\times W\times C}$ and its corresponding ground-truth mask $y_{gt}$.

In the second training phase, an adversarial image~$I_{\text{adv}}$ from~$I_F$ is firstly generated with the Fast Gradient Sign Method~(FGSM)~\cite{goodfellow_arxiv_2014}, a fast and famous adversarial attack method. Due to the fast attack speed of FGSM, we apply it instead of other adversarial attacks as our attack method. Other well-known adversarial attack algorithms, such as BIM~\cite{kurakin_arxiv_2016} and MI-FGSM~\cite{dong_cvpr_2018}, spend more time on attacking an image. Specifically, BIM takes 0.94 second, MI-FGSM takes 1.13 second, while FGSM only take 0.2 second to perform adversarial attack on an image. FGSM attacks the latest gradients of our model to generate an adversarial example~$I_{\text{adv}}\in\mathbb{R}^{H\times W\times C}$, which can be formulated as follows:
\begin{equation}
I_{\text{adv}} = I_F + \epsilon \cdot \text{sign}(\nabla_{I_F}L(\theta,I_F,y_{gt})),
\label{eq:fgsm}
\end{equation}
where~$\theta$ denotes the current parameters of our model,$\nabla_{I_F}$ denotes obtaining the gradients of our model when input $I_F$ and~$L$ denotes the loss function. $\epsilon$ is taken a random number in the range $(0,0.01]$ in every iteration to increase randomness, which generates more training data during SAT and would make the network more robust. The model is then trained on the obtained adversarial example~$I_{\text{adv}}$ and $y_[gt]$, the corresponding ground-truth mask also used in the first training phase. We update the parameters $\theta$ of our network in Step 6 and Step 14 by back-propagation to minimize the loss function of Eq.~\ref{eq:loss}.

\begin{figure}[t]
	\centering
	\includegraphics[width=0.5\textwidth]{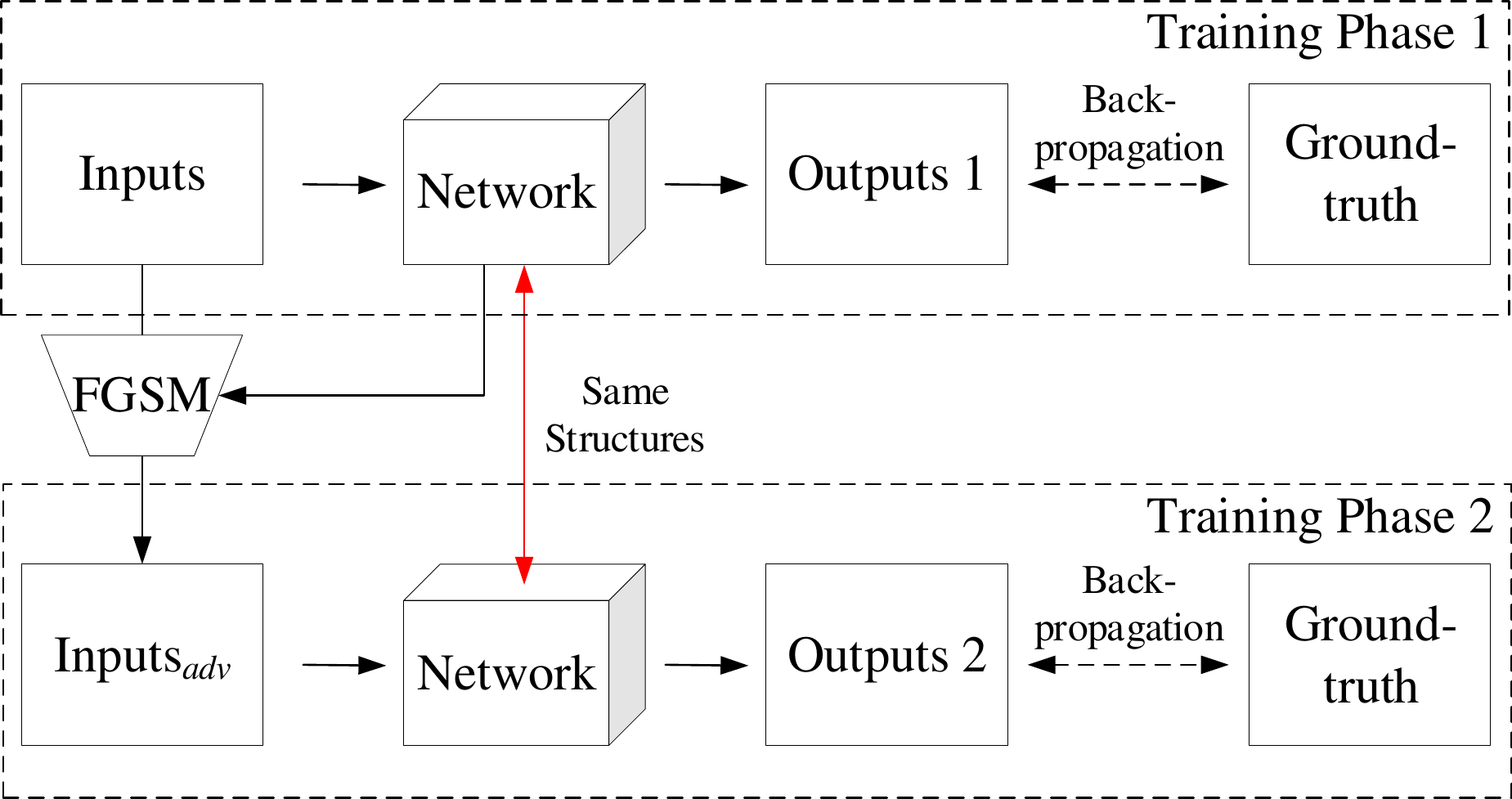}
	\caption[]{The flowchart of our training phase.}
	\label{flowchat_3_1}  
\end{figure}

We draw a flowchart to explain the training phase in
Fig.~\ref{flowchat_3_1}. The first
phase is the traditional training process. In a single
training iteration, we add a second phase to perform
SAT. Specifically, after the normal back-propagation using the
input-ground-truth pairs and updating the network's
parameters, the same inputs are used to generate adversarial
samples $Input_{adv}$ by adversarial attack method, FGSM,
according to the updated model. Although $Input_{adv}$ has the
same content as $Input$, which can be localized the tampered
regions by the updated model, $Input_{adv}$ is with
adversarial samples and can mislead the detection result,
shown in Fig.~\ref{sat}(b). In other
words, $Input_{adv}$ is new data for the model. This phase
increases the training data dynamically since the adversarial
samples are generated according to the updated model's
parameters in every training iteration. Note that we optimize
the same network in both the first and the second training
phase.


With two-phase self-adversarial training, SAT can provide new training data dynamically from limited original samples. Those training data generated via adversarial attacks, e.g. FGSM, can make our model more robust.


To illustrate the impacts of SAT, we draw the residual map between $I_F$ and~$I_{\text{adv}}$ and then enlarge it by 20 times, as shown it in Fig.~\ref{sat}(a). The residual map will change in the following epoch because~$I_{\text{adv}}$ is constantly changing. Meanwhile, to differ the forged image~$I_F$ and its corresponding adversarial example~$I_{\text{adv}}$, we illustrate the inference results of them before training~$I_{\text{adv}}$ in Fig.~\ref{sat}(b). As shown in Fig.~\ref{sat}(b), before the second training phase, the model predicts an approximately precise mask using~$I_F$ but predicts a wrong mask using~$I_{\text{adv}}$. 

\subsection{Visualization Analysis}
\label{sec:visual}
To explore the internal mechanism of our proposed CW-HPF and \textit{\textbf{forgery attention}}, visualization analysis was conducted in this section. As illustrated in Fig.~\ref{hpf_c}, a forged image was taken as examples. We visualized the activation maps of different noise features extracted from a normal high pass filter layer~\cite{zhou_cvpr_2018} and our proposed CW-HPF. CW-HPF uses the inter-channel information while HPF does not. Then in Figs.~\ref{fa_c0} and~\ref{fa_c}, we visualized the activation maps of different attention feature maps predicted by a normal attention module~\cite{fu_cvpr_2019} and our proposed \textit{\textbf{forgery attention}}. The activation maps with heatmaps were super-imposed on the forged image, where the red regions were with high activation values while the blue ones were with low values.
\subsubsection{CW-HPF}
\begin{figure}
	\centering
	\includegraphics[width=.5\textwidth]{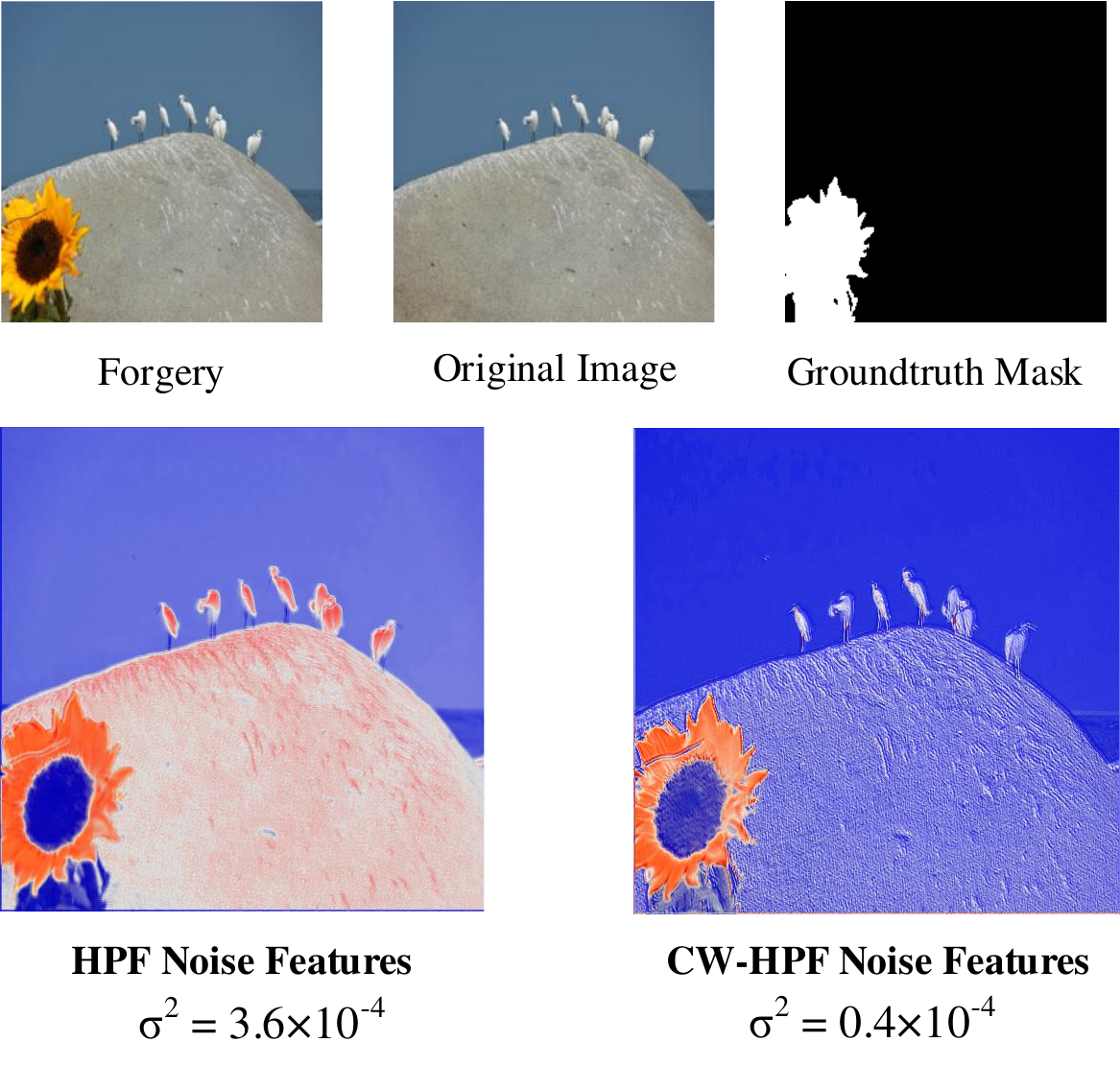}
	\caption{Visualization of different noise features extracted from an example image~(NIST2016\_4867.jpg) from NIST. From the visualization we can observe that HPF noise features are highly responsive at rock and birds in red, which are pristine regions, while ours are not. It indicates that CW-HPF noise features respond weakly to pristine regions but more intensive on tampered regions.}
	\label{hpf_c}
\end{figure}
As we can see in Fig.~\ref{hpf_c}, the flower (fake),
the rock (pristine) and the birds (pristine) all naturally
contain high-frequency information, which both normal HPF
layer and our proposed CW-HPF can extract. However, we can see
HPF extracts all high-frequency information while our proposed
CW-HPF can just focus on the more precise noise features in
the manipulated area by leveraging the inter-channel
relationships. As a result, the flower is much more prominent
in the heatmap corresponding to our proposed CW-HPF.

The variance of each noise feature map was also calculated. As shown in Fig.~\ref{hpf_c}, the variance~$\sigma^2$ of our CW-HPF noise features is only one-ninth of that of HPF noise features, which also reveals that CW-HPF noise features have a smaller internal gap. The small variance enables CW-HPF to extract more consistent noise features in the tampered regions.

One possible reason of this phenomenon is that our CW-HPF takes advantage of inter-channel relationships to enhance the correlation between all channels. These relationships amplify the slight perturbations of noise inconsistency. Therefore CW-HPF focuses more on the tampered regions.

\subsubsection{Forgery Attention}

\begin{figure}
	\centering
	\includegraphics[width=.5\textwidth]{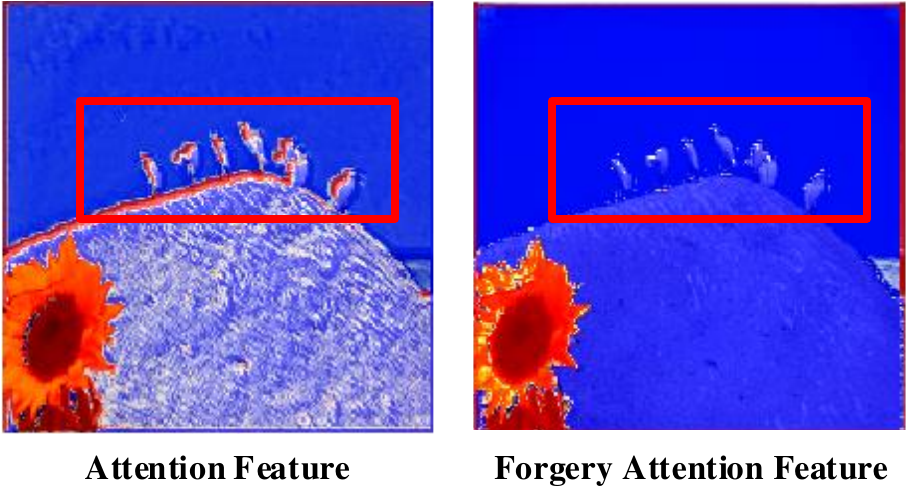}
	\caption{Visualization of different attention features of the example image in Fig.~\ref{hpf_c}. The red box emphasizes untampered objects, namely the birds. The untampered area of the left heatmap is redder than that of the right one, which indicates that the normal attention module responds to all objective regions while our \textit{\textbf{forgery attention}} focuses more on the tampered regions.}
	\label{fa_c0}
\end{figure}

\begin{figure}
	\centering
	
	\includegraphics[width=.5\textwidth]{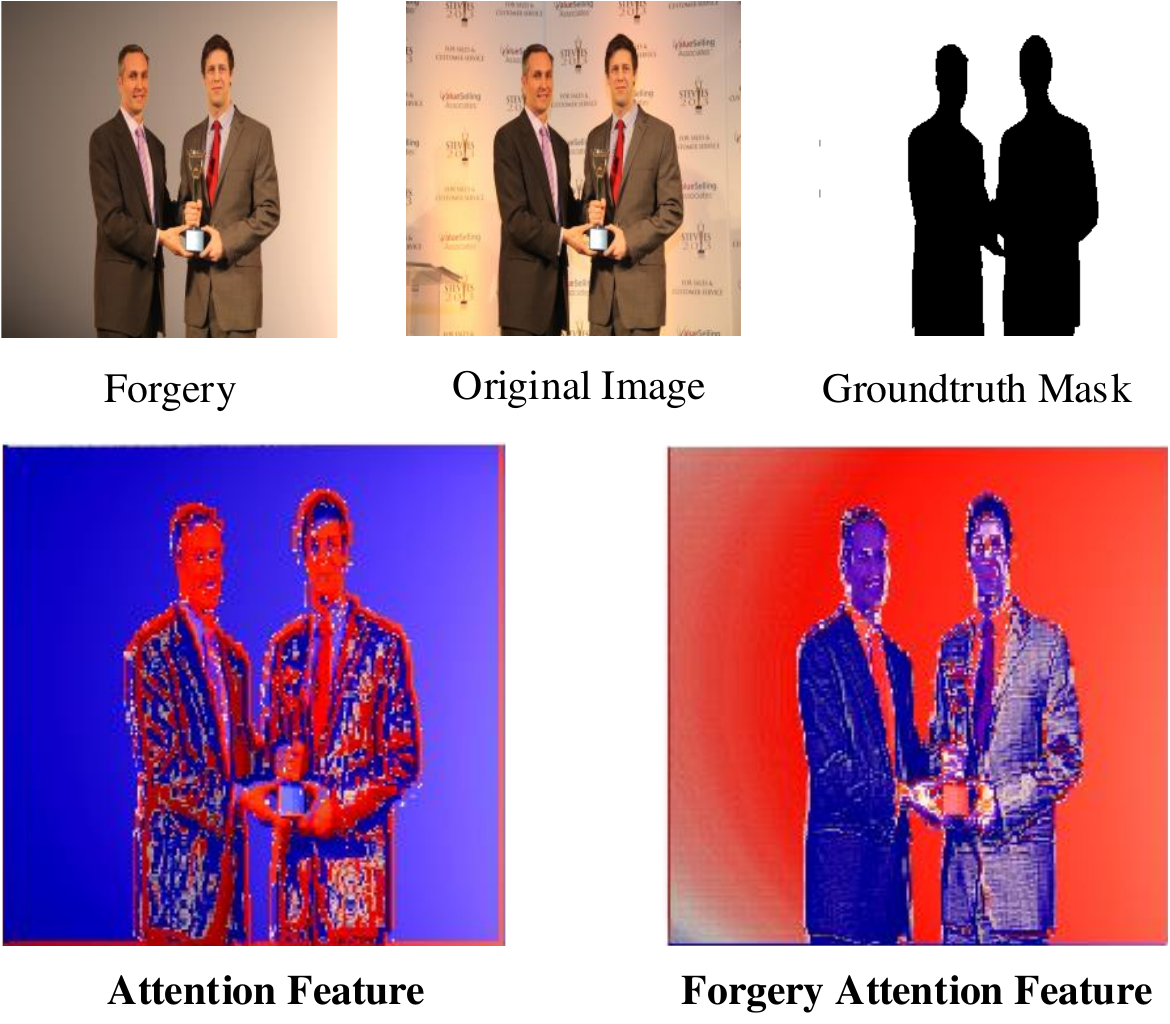}
	\caption{Visualization of different attention features of an example image (NIST2016\_3119.jpg) from NIST 2016. The normal attention module responds to the objects (i.e., persons) in red while our \textit{\textbf{forgery attention}} responds more to the tampered regions (the background).}
	\label{fa_c}
\end{figure}

Figs.~\ref{fa_c0} and~\ref{fa_c} visualize different regional attention of normal attention features and our proposed \textbf{\textit{forgery attention}} features. Especially in Fig.~\ref{fa_c}, the redder regions in the heatmap of normal attention features fell in two persons while \textit{\textbf{forgery attention}} had higher responses to the background. It indicates that the \textit{\textbf{forgery attention}} focuses more on the tampered regions instead of the salient objects. That's because architectures of normal attention modules are designed to detect the texture of objects. On the contrary, our \textit{\textbf{forgery attention}} aims to pay more attention to the tampered traces.

The possible reason why \textit{\textbf{forgery attention}} aims to pay more attention to the tampered traces is that it explores global contextual noise information by building associations among noise features with the attention mechanism. In addition, it can also be attributed to CW-HPF which extracts the noise features first. Our method can adaptively aggregate long-term contextual information of noise features, thus improves feature representation for forgery localization. 

Specifically, two attention branches contribute to focusing on noise interdependencies. It can be inferred from Equation~\ref{saf} that the resulting feature~$\text{SAF}_N$ at each position is a weighted sum of the noise features across all pixels and the original feature, which gives it a global contextual view and selectively aggregative contexts. Similar noise features achieve mutual gains, thus improving the noise consistency between the tampered and original regions. Furthermore, Equation~\ref{caf} shows that the final feature of each channel is a weighted sum of the noise features of all channels and original features, which further models long-term noise dependencies among feature maps to boost noise feature discriminability. Each high-level channel can be regarded as a class-specific response, and different noise responses are associated with each other. Exploiting the interdependencies between channel maps emphasizes interdependent feature maps and improves the feature representation of tampered traces. Therefore, \textit{\textbf{forgery attention}} achieves better performance in forgery localization tasks.


\section{Experiments}
\label{sec:experiment}
In this section, we carry out comprehensive experiments to demonstrate
our proposed approach on several benchmark datasets and compare the
results with state-of-the-art methods. Besides, we evaluate the
robustness of our method in the scenarios of resizing, JPEG
compression, and adversarial attacks. All of our
experimental results are obtained on real-world datasets under the
actual hardware experiments. There are no numerical simulations in our
experiments. The source codes and auxiliary
materials are available for download from
GitHub~\footnote{\url{https://github.com/tansq/SATFL}}.

\subsection{Setup}

\begin{table*}
	
	\caption{Quantitative results of ablation studies on DEFACTO.}
	\label{model_ablation}
	\centering
	\resizebox{0.7\textwidth}{!}{
		\begin{tabular}{ccc}
			
			\toprule
			Model                   & AUC   & $F_1$    \\
			\midrule
			Baseline                & 0.978 & 0.855 \\
			
			CW-HPF Model            & 0.986 & 0.878 \\
			
			\textit{\textbf{Forgery Attention}} Model                & 0.990 & 0.890 \\
			\midrule
			Coarse-to-fine \textit{\textbf{Forgery Attention}} Model & 0.992 & 0.904 \\
			Coarse-to-fine \textit{\textbf{Forgery Attention}} Model+SAT & 0.996 & 0.920 \\
			\midrule
			Coarse-to-fine \textit{\textbf{Forgery Attention}} Model+SAT+Flipping+Rotation & \textbf{0.998} &\textbf{ 0.929 } \\
			\bottomrule
		\end{tabular}
	}
\end{table*}

\subsubsection{Datasets}
The following datasets are used in our experiments:
\begin{itemize}

\item 
\textbf{DEFACTO}~\cite{mahfoudi_espc_2019} is a synthesized dataset generated from MSCOCO~\cite{lin_eccv_2014}. Three typical types of forgeries (i.e., splicing, copy-move, and removal) are involved in DEFACTO. 98,779 tampered images are selected from DEFACTO as our base dataset for ablation study and pre-training. We have to emphasize that our base dataset contains fewer images than other studies', such as Mantra-Net~\cite{wu_cvpr_2019} (102,028 samples) and SPAN~\cite{hu_eccv_2020} (102,028 samples). The training-testing ratio is set to 9:1.

\item 
\textbf{Columbia}~\cite{ng_cucdl_2009} provides 180 splicing images with edge masks. The ground-truth masks are generated by ourselves from the corresponding edge masks
\item 
\textbf{CASIA}~\cite{dong_csicsip_2013} contains splicing and copy-move images in which forged regions are carefully selected. Some forged images have been further post-processed with filtering and blurring. It can be split into CASIA 2.0 (5,123 samples) for training and CASIA 1.0 (921 samples) for testing. Both of them are provided with ground-truth masks.

\item 
\textbf{COVERAGE}~\cite{wen_icip_2016} contains 100 forged images manipulated by copy-moving. All the images are post-processed to remove visual traces. It is provided with ground-truth masks.

\item
\textbf{NIST}~\cite{guan_wacvw_2019} is composed of 564 samples manipulated with splicing, copy-move, or removal. The visible traces of manipulations are concealed by post-processing. The dataset has ground-truth masks for evaluation.

\item 
\textbf{PS-dataset} is a large-resolution dataset introduced in the latest work~\cite{zhuang_tifs_2021}. It involves three sub-datasets, namely PS-boundary dataset, PS-arbitrary dataset and PS-script dataset. Among these sub-datasets, the PS-boundary dataset and PS-arbitrary dataset are created manually with Photoshop$^\circledR$, while the PS-script dataset is tampered with the automatic script.

\end{itemize}

Please note that for the sake of fair comparison, our experiments follow the training-testing ratio configuration in RGB-N~\cite{zhou_cvpr_2018} on NIST, COVERAGE, and CASIA.

\subsubsection{Implementation Details}
Our approach has been implemented based on TensorFlow. Our framework was trained with input images resized to~$512\times 512$ on a single Tesla P100 GPU.\footnote{OpenCV package was used to resize the images. We set interpolation=INTER\_AREA when resizing the forged images and set interpolation=INTER\_NEAREST for resizing the masks.} An ADAM solver was used to optimize the model with a learning rate of 0.002.

\subsubsection{Evaluation Metrics}
Pixel-level $F_1$ score and Area Under the receiver operating Curve
(AUC) were employed as our evaluation metrics. $F_1$ score and AUC
measured the performance of binary classification for every pixel,
where higher scores indicate better performance. Both pixel-level AUC
and $F_1$ score values range in [0,1]. Kindly note according to
observations based on our experimental results, $F_1$ score is more
accurate than AUC. This is due to the fact that $F_1$ score drops
significantly when the predicted masks contain quite a few
false-positive predictions~(the pristine pixels are marked as
tampered). On the contrary, AUC still remains at a high score on
account of some factors such as threshold settings in this case.

\subsubsection{Loss Function}
During training, binary cross-entropy loss was used as our training loss function. It was minimized by optimizing the model parameters. The loss function is formulated as follows:
\begin{equation}
 	\text{Loss} = L_{\text{BCE}}(y_{\text{gt}},y_{M_1})+L_{\text{BCE}}(y_{\text{gt}},y_{M}),
 	\label{eq:loss}
\end{equation}
where $L_{\text{BCE}}$ denotes Binary Cross-Entropy loss, ~$y_{\text{gt}}$ denotes ground-truth masks, $y_{M_1}$ denotes coarse masks, and~$y_M$ denotes refined masks.

\subsubsection{Compared Methods}
We have selected several advanced methods compared with our approach. The advanced methods can be categorized into two types, namely unsupervised methods and deep-learning based methods. The unsupervised methods~\cite{krawetz_hfs_2007,mahdian_ivc_2009,ferrara_tifs_2012} leverage feature extraction techniques to seek unnatural traces. Recently, the deep-learning based methods~\cite{bappy_tip_2019,zhou_cvpr_2018,wu_cvpr_2019,hu_eccv_2020} have demonstrated superior performance by using convolutional neural networks. In our work, we have compared our approach with these two types of methods.

\subsection{Ablation Study}

We have conducted extensive ablative experiments to validate each component of our framework. In particular, we have first generally conducted the progressive ablation study and then carefully conducted detailed experiments for each of the proposed components. All of the ablative experiments were conducted on the DEFACTO dataset. 

\subsubsection{Progressive Ablation Study}

The progressive ablative experiments were conducted to validate our proposed components and SAT. The setup models were as follows:
\begin{itemize}
	\item \textbf{Baseline:} The baseline model contained an HPF filter layer, a dilated convolutional module, and VGG blocks. It did not have a coarse-to-fine architecture.
	\item \textbf{CW-HPF Model:} The HPF filter layer was replaced in the baseline model with a CW-HPF filter block. Others remained the same.
	\item \textbf{\textit{\textbf{Forgery Attention}} Model:} It was constructed based on the CW-HPF model, composed of CW-HPF, VGG blocks, a dilated convolutional module, and a \textit{\textbf{forgery attention}} module.
	\item \textbf{Coarse-to-fine Forgery Attention Model:} A coarse-to-fine net was constructed using the CW-HPF model as the \textit{coarse net} and \textit{\textbf{forgery attention}} model as the \textit{refined net}. This is actual our proposed model.

\end{itemize}

All models were trained using the same setting, and the results are reported in Table~\ref{model_ablation}. From Table~\ref{model_ablation}, it can be seen that our proposed components are all effective and improve the AUC and $F_1$ scores significantly. The CW-HPF block improves the performance by 0.008 in AUC and 0.023 in $F_1$. The \textit{\textbf{forgery attention}} module further improves by 0.004 in AUC and 0.012 in $F_1$ over the CW-HPF model. The coarse-to-fine architecture further improves by 0.002 in AUC and 0.014 in $F_1$ over the \textit{\textbf{forgery attention}} model. 

To evaluate our proposed self-adversarial training strategy, data augmentation techniques were applied for the well-trained coarse-to-fine forgery attention model, namely our proposed SAT plus flipping and rotation. As also shown in Table~\ref{model_ablation}, our proposed SAT further boosts the performance by 0.004 in AUC and 0.016 in $F_1$. SAT provides dynamic training data for more robust performance. For optimal performance, flipping and rotation were employed to further augment the training dataset and achieve 0.998 in AUC and 0.929 in $F_1$.

	\subsubsection{Ablation Study for Forgery Attention}
	We have applied four models that
	utilize different attention structures on top of basic CW-HPF
	model: 
	\begin{itemize}
		\item \textbf{CAM only}: Only CAM was used as attention module; 
		
		\item \textbf{PAM only}: Only PAM was used as attention module;
		
		\item \textbf{Dual Attention (Softmax)}: CAM and PAM were combined
		in parallel but with Softmax layer instead of Sigmoid
		layer.
		
		\item \textbf{Dual Attention (w/o CW-HPF)}: An alternative dual
		attention~\cite{fu_cvpr_2019} module was used without CW-HPF
		block;
		
		\item \textbf{Forgery Attention Model}: Our proposed forgery
		attention module.
	\end{itemize}

	As shown in Table~\ref{attention_ablation}, a single CAM or a single PAM slightly decreases the performance of the basic model. Combining CAM and PAM makes good use of spatial and channel dependencies to generate more precise results. Adopting Softmax instead of Sigmoid, results in
    decreased AUC and $F_1$ as well. Meanwhile, applying the common dual attention
    module~\cite{fu_cvpr_2019} without CW-HPF cannot
    improve the performance. This ablative study clearly shows that our proposed attention architecture with dual attention module, Sigmoid layer and
    CW-HPF is more adaptive to forgery localization task. Three
    intuitive reasons can be concluded for the effectiveness of
    our proposed dual attention module. Firstly, spatial and
    channel-wise contextual dependencies both are important to
    distinguish the intrinsic inconsistency; Secondly, Softmax,
    which is often used as the last activation function of a
    neural network, is not suitable to replace Sigmoid in the
    intermediate layers of a deep-learning framework; Thirdly,
    CW-HPF can further provide richer noise features for attention
    modules, which contributes to focusing on the high-pass
    inconsistency between the pristine regions and tampered
    regions.

\begin{table}
	
	\caption{Quantitative results of ablation studies for forgery attention structures on DEFACTO.}
	\label{attention_ablation}
	\centering
	\resizebox{0.4\textwidth}{!}{
	\begin{tabular}{ccc}
		
		\toprule
		Model                   & AUC   & $F_1$    \\
		\midrule
		CW-HPF Model            & 0.986 & 0.878\\
		\midrule
		CAM only                & 0.962 & 0.840 \\
		PAM only                & 0.984 & 0.856 \\
		Dual Attention (Softmax)            & 0.986 & 0.879 \\
		Dual Attention (w/o CW-HPF)            & 0.985 & 0.879 \\
		\midrule
		\textit{\textbf{Forgery Attention}} Model                &\textbf{ 0.990} & \textbf{0.890} \\
		
		\bottomrule
	\end{tabular}
}
\end{table}

\subsubsection{Ablation Study for SAT}

SAT is proposed to alleviate the problem of limited training data. SAT exploits adversarial attacks in every training iteration and generates new training data dynamically, which guides our model to defend from adversarial attacks and achieve more robust performance. Unlike traditional data augmentation, our SAT can provide unlimited new adversarial training data according to model updating. The main difference between SAT and regular data augmentation, such as flipping, is that SAT is based on the model's parameters while others focus on the given data. We propose SAT to explore the possibility of augmenting data through model parameters instead of through data. Traditional data augmentation techniques do not conflict with our SAT but can further improve the performance with SAT. However, the common augmentation would not benefit for defense against the attacks. A simple augmentation flipping has been used in training our network on the DEFACTO dataset. As seen in Table~\ref{sat_flip_3_6}, both SAT and Flipping + Rotation improved the performance of the baseline model. Based on the model with flipping and rotation, SAT can further increase the detection scores by 0.004 in AUC and 0.004 in $F_1$. However, SAT enables the network to decrease slightly while flipping and rotation do not when the forged images are imposed with Gaussian Noise attack. Specifically, the network trained with Flipping + Rotation degrades by 17\% in AUC and 18\% in $F_1$ after attack while the network trained with SAT only decreases by 8\% in AUC and 10\% in $F_1$. Since the adversarial attack used in SAT is a powerful attack technique, the network can be more robust when meeting other attacks.

\begin{table}[t]
	
	\caption{Quantitative results of flipping+rotation and SAT on DEFACTO dataset.}
	\label{sat_flip_3_6}
	\centering
	\resizebox{0.44\textwidth}{!}{
		\begin{tabular}{ccc}
			
			\toprule
			Data Augmentation             & AUC   & $F_1$     \\
			\midrule
			Baseline & 0.992&0.904 \\
			\midrule
			Flipping + Rotation           & 0.993 & 0.921  \\
			After Gaussian Noise Attack & 0.819   &  0.757  \\
			
			\midrule
			SAT          & 0.996 & 0.920 \\
			
			After Gaussian Noise Attack &  0.915  &  0.826  \\
			
			\midrule
			Flipping + Rotation + SAT          & 0.998 & 0.929 \\
			
			After Gaussian Noise Attack &  0.937  &  0.903  \\
			\bottomrule
		\end{tabular}
	}
\end{table}

\subsubsection{Ablation Study for CW-HPF}
The ablative experiments have been conducted
	to demonstrate the effectiveness of CW-HPF with parallel spatial and
	channel attention modules. Specifically, seven different feature
	extraction strategies have been evaluated in our proposed
	coarse-to-fine forgery attention module as follows:
\begin{itemize}
		\item \textbf{RGB + RGB}: there was no HPF layer in neither the coarse net nor the refined net; 
	
	\item \textbf{HPF + RGB}: there was only a normal HPF
	layer~\cite{zhou_cvpr_2018} at the bottom of the coarse net;
	
	\item \textbf{RGB + HPF}: there was only a normal HPF layer at the bottom
	of the refined net;
	
	\item \textbf{HPF + HPF}: the normal HPF layers were adopted at the bottom of
	both the coarse net and the refined net;
	
	\item \textbf{CW-HPF + HPF}: it contained a CW-HPF block at the bottom of
	the coarse net and a normal HPF layer at the bottom of the refined
	net.
	
	\item \textbf{HPF + CW-HPF}: it contained a normal HPF layer at the bottom
	of the coarse net and a CW-HPF block at the bottom of the
	refined net;

	\item \textbf{CW-HPF + CW-HPF}: the CW-HPF blocks were adopted at the bottom of
	both the coarse net and the refined net.
\end{itemize}

\begin{table}[t]
	
	\caption{Quantitative results of ablation studies for CW-HPF structures on DEFACTO.}
	\label{hpf_ablation}
	\centering
	\resizebox{0.36\textwidth}{!}{
	\begin{tabular}{ccc}
		
		\toprule
		Model                   & AUC   & $F_1$    \\
		\midrule
		RGB + RGB                & 0.804 & 0.721 \\
		HPF + RGB                & 0.970 & 0.850 \\
		RGB + HPF                & 0.964 & 0.844 \\
		HPF + HPF                & 0.977 & 0.856 \\
		CW-HPF + HPF                & 0.981 & 0.867 \\
		HPF + CW-HPF          & 0.984 & 0.869 \\
		
		\midrule
		CW-HPF + CW-HPF              &\textbf{0.992} & \textbf{ 0.904} \\
		
		\bottomrule
	\end{tabular}
}
\end{table}

As we can see in Table.~\ref{hpf_ablation}, the proposed method with two CW-HPF modules outperforms other methods. Specifically, the method without a high-pass filters layer performs poorly in final results, with only 0.804 in AUC and 0.721 in $F_1$. Applying one HPF module in either the coarse net or the refined net improves the performance by 0.160-0.166 in AUC and 0.123-0.129 in $F_1$. Introducing HPF in both the coarse net and the refined net improves the performance in a clear margin by 0.174 in AUC and 0.135 in $F_1$ compared to the model with RGB + RGB. One possible reason is that the tampering traces are hidden in the high-frequency domain, and the high-pass filters can detect them. Then, the model with a CW-HPF block in either the coarse net or the refined net boosts the performance of two HPF blocks by 0.004-0.006 in AUC and 0.011-0.013 in $F_1$. Furthermore, our proposed model with two CW-HPF further increases the detection scores by 0.008 in AUC and 0.035 in $F_1$ compared to the model with a CW-HPF in the refined net and achieves the best performance. It indicates that our proposed CW-HPF module enhances noise features and boosts performance in both the coarse net and the refined net.

\subsubsection{Ablation Study for Dilated Convolutional Module}

DFCN has used a series of dilated convolution layers to enlarge receptive fields to avoid learning features from narrow local regions. Thus following their approach, four dilated convolution layers are applied to inflate the kernels by inserting zeros between kernel elements with different dilation rates for extracting features with larger receptive fields. We follow DFCN's settings, and the dilation rates in the four dilated convolution layers are 2, 4, 8, and 16, respectively. The dilated convolution layers bridge the encoder and the corresponding decoder. There are two sub-nets in a coarse-to-fine manner. Here, we have conducted ablative experiments to validate the effectiveness of the dilated convolutional module. 

To compare with our final model, the coarse-to-fine forgery attention model of the revision, with two dilated convolutional modules in both the coarse net and the refined net, we have modified the bridge of the encoders and the decoders of two sub-nets as follows:
\begin{itemize}
	\item \textbf{Dilated + Dilated}: it contained two dilated convolutional modules in the coarse net and the refined net.
	
	\item \textbf{w/o Dilated + w/o Dilated}: it directly connected the encoders and decoders without any dilated convolutional module.
	
	\item \textbf{Dilated + w/o Dilated}: it included a dilated convolutional module in the coarse net and no dilated convolutional module in the refined net.
	
	\item \textbf{w/o Dilated + Dilated}: it consisted of a dilated convolutional module in the refined net and no dilated convolutional module in the coarse net.
\end{itemize}

\begin{table}[t]
	
	\caption{Quantitative results of different settings of dilated convolutional module on DEFACTO dataset.}
	\label{dialted_conv}
	\centering
	\resizebox{0.4\textwidth}{!}{
	\begin{tabular}{ccc}
		
		\toprule
		Model             & AUC   & $F_1$     \\
		\midrule
		Dilated + Dilated &  0.996 & 0.920 \\
		\midrule
		w/o Dilated + w/o Dilated & 0.993   &  0.910 \\
		
		Dilated + w/o Dilated &  0.990  &  0.907  \\
		w/o Dilated + Dilated   & 0.994 & 0.913  \\
		
		\bottomrule
	\end{tabular}
}
\end{table}

As we can see in Table~\ref{dialted_conv}, without any dilated convolutional modules, the model decreases the performance by 0.003 in AUC and 0.010 in $F_1$. Interestingly, if a dilated convolutional module is used in the coarse net, the model's performance declines slightly compared to that without a convolutional module. Meanwhile, the model with a dilated convolutional module in the refined net increases by 0.001 in AUC and 0.003 in $F_1$ compared to the model w/o Dilated + w/o Dilated, but decreases by 0.002 in AUC and 0.007 in $F_1$. It indicates that the coarse-to-fine network is required to generate the richer features in the refined net than in the coarse net, and applying the dilated convolutional modules in both the coarse net and the refined net improves the performance.

\subsubsection{Ablation Study for Coarse-to-Fine Connection}

Generally, referring to the coarse-to-fine manner, the connection from the coarse net to the refined net is to directly transmit the results of the coarse net as the inputs of the refined net. However, the result of the coarse net is a binary mask, which does not contain any semantics for refinement in forgery localization. Therefore, rather than using the final results of the coarse net as the connection, we deliver complete feature information generated by the last deconvolutional layer of the coarse net to the refined net, which enables the refined net to be optimized along with the features. We validate this design through an ablative experiment. We adopt the CW-HPF model as the coarse net and forgery attention model as the refined net and set two types of bridges between the coarse net and the refined net. The first type of connection uses the results of the coarse net, defined as Direct Connection, while the second one applies the output features of the last deconvolutional layer of the coarse net, defined as Feature Connection.

\begin{table}[t]
	
	\caption{Quantitative results of different settings of connection on DEFACTO dataset.}
	\label{connection}
	\centering
	\resizebox{0.36\textwidth}{!}{
	\begin{tabular}{ccc}
		
		\toprule
		Model             & AUC   & $F_1$     \\
		\midrule
		Direct Connection   & 0.709 & 0.322  \\
		Feature Connection &  0.996 & 0.920 \\

		\bottomrule
	\end{tabular}
}
\end{table}

The results are shown in Table~\ref{connection}. The coarse-to-fine manner with Direct Connection shows a poor performance compared to the Feature Connection significantly. Specifically, the AUC of Feature Connection is 40\% higher and the $F_1$ is 185\% than Direct Connection. The possible reason is that Feature Connection provides integral features for the refined net while Direct Connection does not.

\subsubsection{Trade-off Experiments}

We have conducted the trade-off experiments based on the Coarse-to-fine forgery attention model to achieve a better trade-off. We employ different numbers of basic filters (8, 16, 32, 48, 64) to conduct the trade-off experiments on the DEFACTO dataset. The different number of basic filters (nbf) represents the last deconvolutional layer's filter numbers while other filters of convolution layers in the network have multiples. The more filters indicate the larger parameters and cost, whereas the double nbf indicates the double network parameters and four calculation resources. Therefore, we gain our trade-off by this experiment.

\begin{figure}[t]
	\centering
	\includegraphics[width=0.5\textwidth]{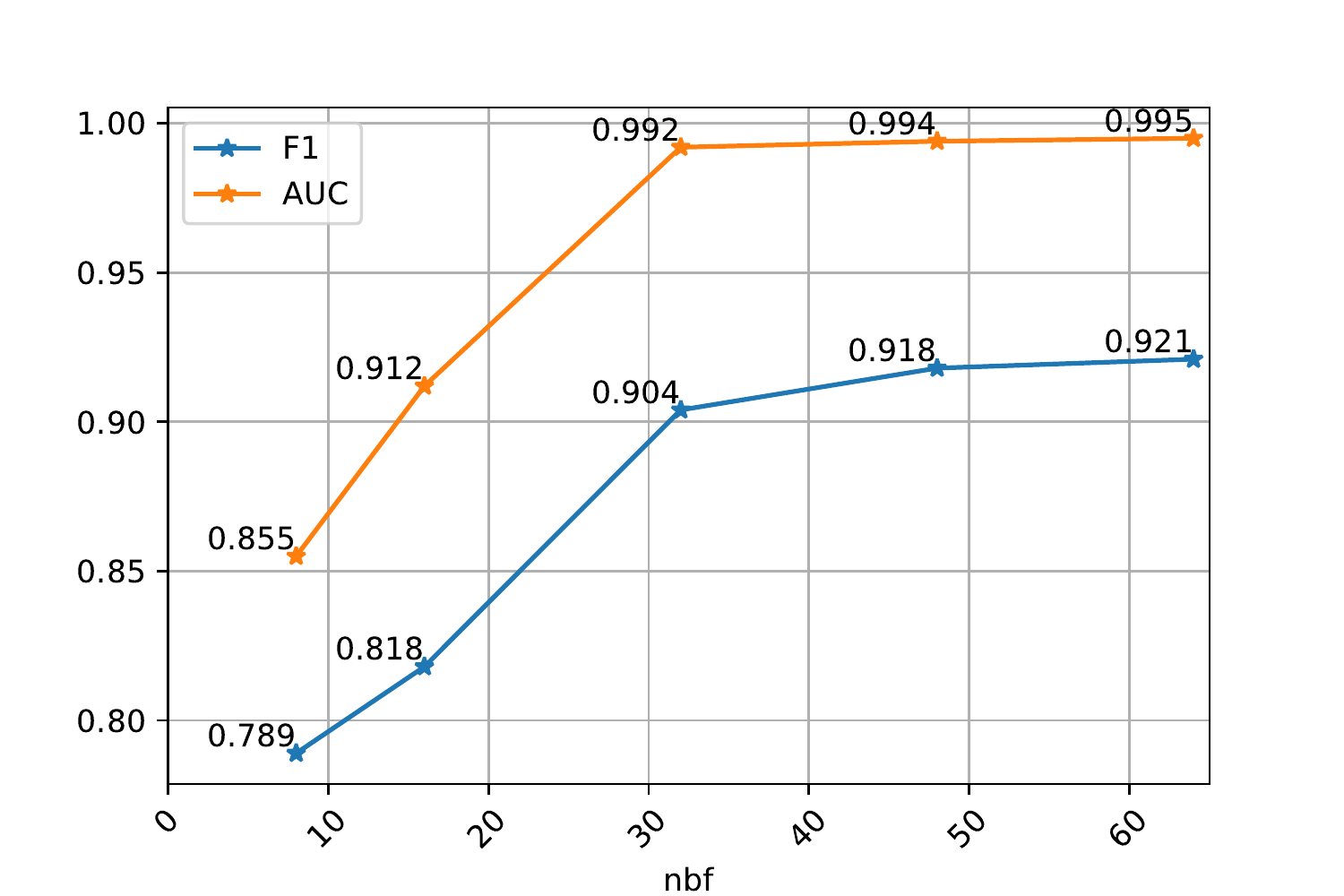}
	\caption[]{The results of trade-off experiments.}
	\label{fig:trade-off}  
\end{figure} 

The results are shown in Fig.~\ref{fig:trade-off}. As we can see, there is a significant rise of $F_1$ score from 8 to 32 nbf while the raising speed slows down dramatically after 32 nbf. We can conclude that we achieve a better trade-off when applying 32 as the number of basic filters.

We have also added this experiment into the main text of the revised version in Section III B. For details, please refer to the corresponding context.

\subsection{Quantitative Results Compared against State-of-the-art Methods}
We compare the performance of our framework against several related methods on four benchmarks, namely NIST, COVERAGE, Columbia, and CASIA. The related methods include classic unsupervised methods, such as ELA~\cite{krawetz_hfs_2007}, NOI1~\cite{mahdian_ivc_2009}, and CFA1~\cite{ferrara_tifs_2012}, and the latest deep networks, including H-LSTM~\cite{bappy_tip_2019}, RGB-N~\cite{zhou_cvpr_2018}, Mantra-Net~\cite{wu_cvpr_2019}, and SPAN~\cite{hu_eccv_2020}. We evaluated our framework under different setups: (1) benchmark training only; (2) fine-tuning. Under the benchmark training only setup, our model was trained on each benchmark separately. Under the fine-tuning setup, to achieve optimal performance, our pre-trained model was fine-tuned using several benchmarks, including NIST, COVERAGE, and CASIA, and tested on the corresponding testing split. Note that a training-testing ratio was set to 7:3 during benchmark training on the Columbia dataset. The actual reason has been given in accompanying discussions of Table~\ref{result}.

\begin{table*}[!t]
	\caption{Quantitative results compared against related methods. --- denotes the result is unavailable using the original method and * denotes we use the different settings to adjust our pre-trained method to Columbia dataset.}
	\label{result}
	\centering
	\resizebox{0.95\textwidth}{!}{
	\begin{tabular}{@{}cccccccccc@{}}
		\toprule
		\multirow{2}{*}{Method} & \multirow{2}{*}{Training Method} & \multicolumn{2}{c}{NIST} & \multicolumn{2}{c}{COVERAGE} & \multicolumn{2}{c}{Columbia} & \multicolumn{2}{c}{CASIA}                                                 \\ \cmidrule(l){3-10}
		&                                  & AUC                      & $F_1$                           & AUC                          & $F_1$                        & AUC       & $F_1$        & AUC       & $F_1$        \\ \midrule
		ELA                     & unsupervised                     & 0.429                    & 0.236                        & 0.583                        & 0.222                     & 0.581     & 0.470     & 0.613     & 0.214     \\
		NOI1                    & unsupervised                     & 0.487                    & 0.285                        & 0.587                        & 0.269                     & 0.546     & 0.574     & 0.612     & 0.263     \\
		CFA1                    & unsupervised                     & 0.501                    & 0.174                        & 0.485                        & 0.190                     & 0.720     & 0.467     & 0.522     & 0.207     \\ \midrule
		H-LSTM                  & fine-tuning                      & 0.794                    & ---                          & 0.712                        & ---                       & ---       & ---       & ---       & ---       \\
		RGB-N                   & fine-tuning                      & 0.937                    & 0.722                        & 0.817                        & 0.437                     & 0.858     & 0.697     & 0.795     & 0.408     \\
		Mantra-Net              & pre-training                     & 0.795                    & ---                          & 0.819                        & ---                       & 0.824     & ---       & 0.817     & ---       \\
		SPAN                    & fine-tuning                      & 0.961                    & 0.582                        & 0.937                        & 0.558                     & 0.936     & 0.815     & 0.838     & 0.382     \\\midrule
		Ours                    & benchmark training               & 0.943                    & 0.622                        & 0.856                        & 0.526                     & 0.917     & 0.891     & 0.788     & 0.384     \\
		Ours                 & finetuning                      & \textbf{0.990}                & \textbf{0.878}                    & \textbf{0.985}                    & \textbf{0.843}                 & \textbf{0.999}* & \textbf{0.983}* &\textbf{0.843} &  \textbf{0.592} \\ 
		\bottomrule
	\end{tabular}
}
\end{table*}

The results are reported in Table~\ref{result}. When adopting the benchmark training setup that uses only a small amount of training data, e.g., 75 forged images from COVERAGE, our model shows superior performance compared to state-of-the-art approaches, especially on the Columbia dataset where our $F_1$ score outperforms all other methods. It indicates that our approach does not rely on large-scale training data to achieve decent performance, which shows the effectiveness of our approach. In the fine-tuning setup, our approach takes good advantage of large-scale training data. It can be concluded that our approach is further boosted by large-scale training data. In particular, our approach outperforms SPAN by 0.296 in $F_1$ score on NIST dataset and 0.048 in AUC on COVERAGE dataset. Note that because all of the forged regions in Columbia dataset are large while the forged regions in DEAFCTO are small ones. Thus, there is a domain gap between Columbia and DEAFCTO. So different from the settings of SPAN and Mantra-Net, our results on Columbia dataset are based on 30\% testing data and the other 70\% data is used for finetuning. It is clear from Table~\ref{result} that our model that is trained on DEFACTO dataset and finetuned on each benchmark datasets achieves state-of-the-art performance. The possible reason is that the proposed components and SAT training strategy work jointly and achieve optimal performance.


Furthermore, we compare the effectiveness of our model with the most recent algorithm, i.e., dense fully convolutional network (DFCN)~\cite{zhuang_tifs_2021}. We followed this setting~\cite{zhuang_tifs_2021} and used 512$\times$512 image patches for training while full images for testing. Our model was trained using PS-script dataset and finetuned using only 10\% forged images of several datasets, respectively, i.e., 100 samples in PS-arbitrary dataset, 100 samples in PS-boundary dataset and 56 samples in NIST dataset. Note that in the setting of ~\cite{zhuang_tifs_2021}, which is different from our experiments on NIST dataset in Table~\ref{result}, NIST dataset is split into 512$\times$ 512 patches. Note that DFCN aims to localize the tampered region with high-resolution while other advanced methods do not. Therefore, for a fair comparison, we make two settings on NIST dataset. When we compare our method with other advanced methods, we apply the typical pre-processing process that resizes the tampered images into $512 \times 512$ resolutions. When we compare our method with DFCN, we follow DFCN's setting and crop the tampered images in NIST dataset into $512 \times 512$ patches.  It can be seen from Table~\ref{com_dfcn} that the performance of our model outperforms DFCN on three datasets. Before fine-tuning in each benchmark dataset, our results are slightly lower than DFCN in AUC on PS-arbitrary and NIST datasets. However, our model has been over DFCN by 0.01 in AUC on PS-boundary dataset. About $F_1$ score, our results are about 25\% higher on PS-boundary dataset and 15\% higher on NIST than DFCN's. After fine-tuning on each dataset, our method has made a good improvement and outperforms DFCN on three datasets. In particular, our results achieve about 10\% in $F_1$ score higher on PS-boundary dataset, slightly higher in AUC and $F_1$ score on PS-arbitrary dataset, and about 6\% in AUC on NIST dataset than DFCN. 

\begin{table}
	\caption{Quantitative results against the most recent algorithm, DFCN. Our algorithm was finetuned using 100 samples in PS-arbitrary dataset, 100 samples in PS-boundary dataset and 56 samples in NIST dataset, respectively. }
	\label{com_dfcn}
	
	\resizebox{0.49\textwidth}{0.07\textheight}{
	\begin{tabular}{ccccccc}
		\toprule
		\multirow{2}{*}{Method} &\multicolumn{2}{c}{PS-boundary} &  \multicolumn{2}{c}{PS-arbitrary} & \multicolumn{2}{c}{NIST} \\ 
		\cmidrule(l){2-7} 
		& AUC         & $F_1$      & AUC             & $F_1$          & AUC            & $F_1$          \\ 
		\midrule
		DFCN (w/o fine-tuning)                   & 0.90       & 0.61        & \textbf{0.91}           & 0.57            & \textbf{0.63}           & 0.20           \\
		
		Ours (w/o fine-tuning)                 & \textbf{0.91}      & \textbf{0.76}        &  0.90       & \textbf{0.58}           & 0.61           &\textbf{0.23}   \\
		\midrule
		DFCN (fine-tuning)                   & 0.99       & 0.82        & 0.97           & 0.67            & 0.80           & 0.38           \\
		
		Ours (fine-tuning)                & 0.99       & \textbf{0.90}        & \textbf{0.98}           &\textbf{0.69}         & \textbf{0.85}     & \textbf{0.40} \\

		\bottomrule      
	\end{tabular}
	}
\end{table}

\begin{table}[!t]
	\centering
	\caption{Computational complexity compared with state-of-the-art
		methods. * denotes the training time of one epoch on NIST dataset. ``M''
		represents million, ``ms'' represents milliseconds and ``mins'' represent
		minutes.}
	\label{tab:complexity}
	\resizebox{0.49\textwidth}{0.048\textheight}{
		\begin{tabular}{lcccc}
			\toprule
			Method     & Params(M) & FLOPs(M) & Inference time(ms)  & Training time (mins)* \\
			\midrule
			Mantra-Net & 3.80          & 7.58     & 392    &  16         \\
			SPAN       & 4.06          & 8.11     & 527  &  18         \\
			\midrule
			Our        & 12.31         & 31.26    & 126  & 8             \\
			\bottomrule
			
		\end{tabular}
	}
\end{table}

\begin{table*}
	\centering
	\caption{Robustness analysis of our framework on the NIST dataset. The results are reported in pixel-lever AUC.}
	\label{tab:robustness}
	\resizebox{0.6\textwidth}{!}{
	\begin{tabular}{cccc}
		\toprule
		Manipulations                & Mantra-Net & SPAN   & Ours \\
		\midrule
		None                         & 0.795     & 0.8395 & 0.990  \\
		\midrule
		Resize (0.78x)               & 0.7743     & 0.8324 & 0.984  \\
		Resize (0.25x)               & 0.7552     & 0.8032 &  0.979 \\
		\midrule
		GaussianBlur (kernel size=3) & 0.7746     & 0.8310 &  0.983    \\
		GaussianBlur (kernel size=5) & 0.7455     & 0.7915 &   0.951 \\
		\midrule
		GaussianNoise (sigma=3)      & 0.6741     & 0.7517 &   0.937   \\
		GaussianNoise (sigma=15)     & 0.5855     & 0.6728 &  0.866 \\
		\midrule
		JPEGCompress (quality=100)   & 0.7791     & 0.8359 & 0.978 \\
		JPEGCompress (quality=50)    & 0.7438     & 0.8068 &   0.938   \\
		\midrule
		FGSM (eps=0.02)              & 0.5058     & 0.5401   &  0.986  \\
		\bottomrule
	\end{tabular}
}
\end{table*}

\subsection{Computational Complexity Analyses}

The computational complexities have been calculated and compared to the existing
benchmarks with 512 $\times$ 512 NIST forged images as input in a
single NVIDIA$^{\textrm{\textregistered}}$
Tesla$^{\textrm{\textregistered}}$ P100 GPU card. As for the
benchmarks, we adopt two popular methods, namely
Mantra-Net~\cite{wu_cvpr_2019} and
SPAN~\cite{hu_eccv_2020}, for comparison.

The analytic report can be found in Table~\ref{tab:complexity}, in
which the inference time is the average over randomly selected 1,000
samples, and the training time if the average over 20
epochs on NIST dataset. 

Please note that for deep-learning image forgery localization models,
average training time as well as inference time with every input
images are the better metrics rather than model parameters and FLOPs,
since quite a few existing approaches have adopted complex
training/inference tricks which cannot be measured with only model
parameters and FLOPs.

From Table~\ref{tab:complexity} we can see that compared with
Mantra-Net and SPAN, our proposed framework consumes much less
training time as well as inference time, though our framework is with
more model parameters and FLOPs. 
Our framework takes only roughly 126ms per image on a single Tesla P100 GPU. This is due to the fact that our
proposed framework is trained and validated in a fully end-to-end
manner, while Mantra-Net and SPAN are with quite a few extra off-model
time-consuming operations/calculations, such as the nested-and-sliding
    window based feature extractor, a large number of matrix operations.               

\subsection{Qualitative Results}
\begin{figure*}
	\centering
	\includegraphics[width=1\textwidth]{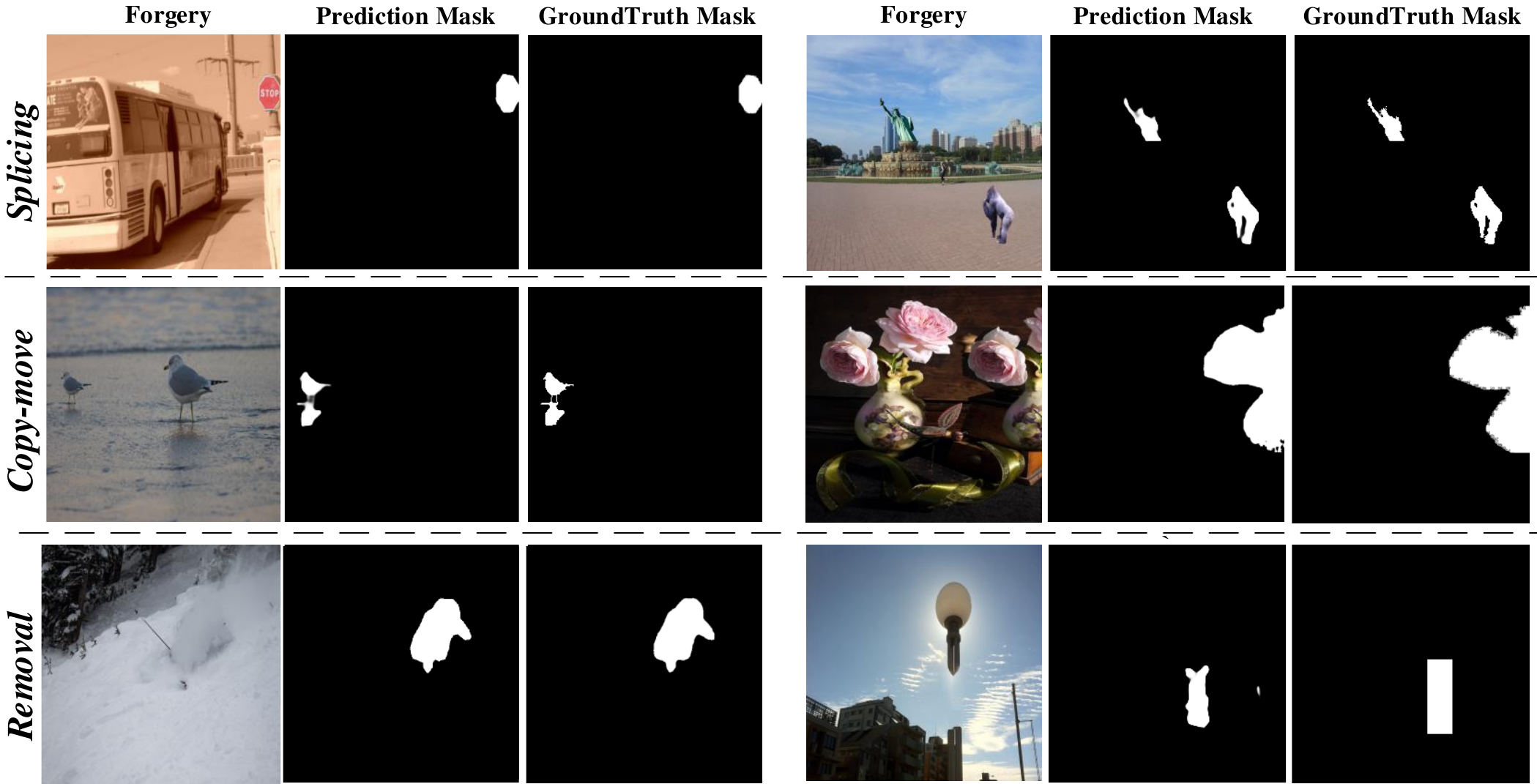}
	\caption{Sample results of our framework in three popular manipulations, namely splicing, copy-move, and removal. The samples are from NIST, Columbia, COVERAGE, CASIA, and DEFACTO.}
	\label{results}
\end{figure*}

In Fig.~\ref{results}, we show the prediction
masks of our proposed framework for some selected images. From
a standalone testing set, we select six tampered images which
are generated with three popular tampering techniques,
including splicing, copy-move and removal, from the mentioned
datasets. As shown in Fig.~\ref{results}, in the tampered
images the original semantics has been changed and
consequently, a considerable understanding gap is caused. For
example, the left splicing image added a stop sign on the
road, damaging the auto-driving system. However, our algorithm
can localize their forged regions credibly. From
Fig.~\ref{results}, it can be seen that our approach produces
accurate results against different tampering techniques. No
matter whether they are tampered objects or background without
recognizable objects such as \textit{snow}, our method detects
them with high precision. In summary, our method makes good
use of spatial and channel-wise attention to noise features
and can precisely spot tampering areas that are obvious or
even indistinguishable to human beings.  

\subsection{Robustness Experiments}

Robustness experiments of our framework have been conducted in this
section. OpenCV built-in functions (including \textbf{AREAResize},
\textbf{GaussianBlur}, \textbf{GaussianNoise}, and
\textbf{JPEGCompress}) and adversarial attacks (\textbf{FGSM}) were
employed to generate content-preserving manipulations on NIST. Note
that epsilons of FGSM used in our SAT strategy were valued from 0 to
0.01 while the epsilon was 0.02 in the testing stage, which is a fair
comparison. As shown in Table~\ref{tab:robustness}, our framework is quite
immune to several types of attacks. All results except FGSM of SPAN
are reported in SPAN~\cite{hu_eccv_2020}.

\section{Conclusion}
\label{sec:conclu}




In this paper, we propose a novel deep neural network solution and a
self-adversarial training strategy to effectively localize tampered
regions in an image.  The major contributions of our work areas
follows: 
\begin{itemize}
\item We have proposed a novel attention mechanism adapting to forgery
  localization task, named \textit{\textbf{forgery attention}} which
  can be used to effectively capture noise feature dependencies in
  both spatial and channel dimensions.
\item We have presented a novel self-adversarial training strategy for
  forgery localization, which augments training data dynamically to
  enable our model to achieve more robust performance, and alleviates
  the problem of limited labeled training data in this scenario.
\item We have proposed a novel forgery localization framework in a
  coarse-to-fine manner, equipped with the Channel-Wise High Pass
  Filter~(CW-HPF) block. Extensive experiments conducted on de-facto
  benchmarking datasets demonstrate that our approach outperforms
  other state-of-the-art solutions in the literature by a clear
  margin.
\end{itemize}

Our future work will mainly focus on two aspects: (1)~introduction of
few-shot learning and even unsupervised learning based strategies to
further tackle the issue of limited training data; (2)~further
exploration of the feasibility of our proposed approach in the wider
multimedia forensics applications, e.g., video forgery localization
and deepfake detection.


%

%
%
%
%
%
%

\ifCLASSOPTIONcaptionsoff
	\newpage
\fi



%

\bibliographystyle{IEEEtran}
\bibliography{reference}

%

%
%
%
\begin{IEEEbiography}
  [{\includegraphics[width=1in,height=1.25in,clip,keepaspectratio]{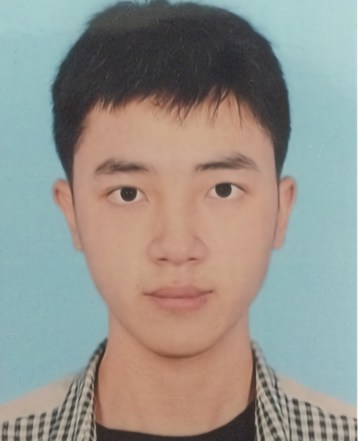}}]{Long
    Zhuo} received the B.S. degree in College of Computer Science and
  Software Engineering from Shenzhen University, Shenzhen, China in
  2019. He is now at Sensetime Group. His current research interests
  include multimedia forensics, image generation and video generation.
\end{IEEEbiography}

\begin{IEEEbiography}[{\includegraphics[width=1in,height=1.25in,clip,keepaspectratio]{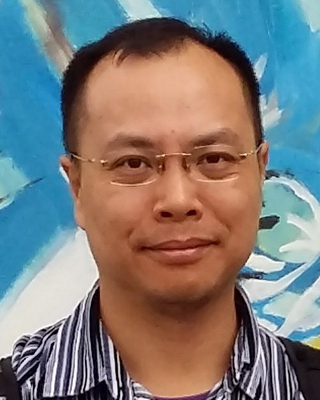}}]{Shunquan Tan (M'10--SM'17)}
  received the B.S. degree in computational mathematics and applied
  software and the Ph.D. degree in computer software and theory from
  Sun Yat-sen University, Guangzhou, China, in 2002 and 2007,
  respectively.

  He was a Visiting Scholar with New Jersey Institute of Technology,
  Newark, NJ, USA, from 2005 to 2006. He is currently an Associate
  Professor with College of Computer Science and Software Engineering,
  Shenzhen University, China, which he joined in 2007. He is the Vice
  Director with the Shenzhen Key Laboratory of Media Security. His
  current research interests include multimedia security, multimedia
  forensics, and machine learning.
\end{IEEEbiography} 

\begin{IEEEbiography}[{\includegraphics[width=1in,height=1.25in,clip,keepaspectratio]{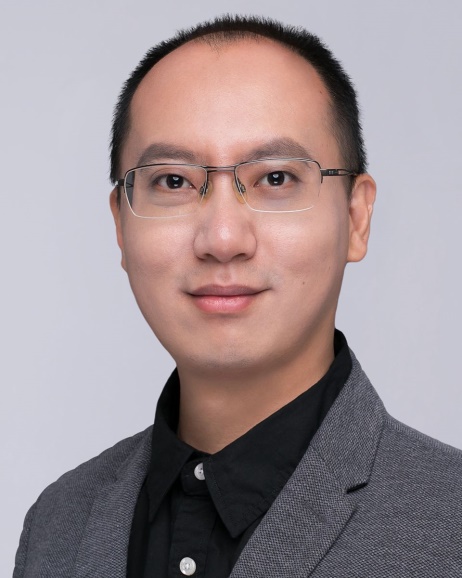}}]{Bin Li (S'07-M'09-SM'17)}
  received the B.E. degree in communication engineering and the Ph.D.
  degree in communication and information system from Sun Yat-sen
  University, Guangzhou, China, in 2004 and 2009, respectively.

  He was a Visiting Scholar with the New Jersey Institute of
  Technology, Newark, NJ, USA, from 2007 to 2008. He is currently a
  Professor with Shenzhen University, Shenzhen, China, where he joined
  in 2009. He is also the Director with the Guangdong Key Lab of
  Intelligent Information Processing and the Director with the
  Shenzhen Key Laboratory of Media Security. He is an Associate Editor
  of the IEEE Transactions on Information Forensics and Security. His
  current research interests include multimedia forensics, image
  processing, and deep machine learning.
\end{IEEEbiography}

\begin{IEEEbiography}[{\includegraphics[width=1in,height=1.25in,clip,keepaspectratio]{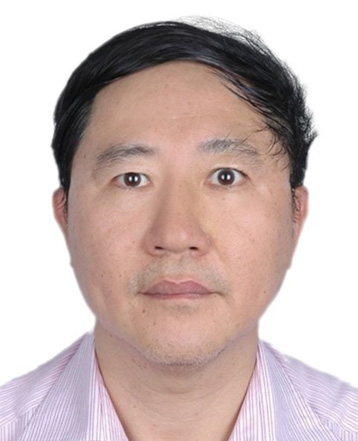}}]{Jiwu Huang (M'98--SM'00--F'16) }
  received the B.S. degree from Xidian University, Xi’an, China, in
  1982, the M.S. degree from Tsinghua University, Beijing, China, in
  1987, and the Ph.D. degree from the Institute of Automation, Chinese
  Academy of Science, Beijing, in 1998. He is currently a Professor
  with the College of Electronics and Information Engineering,
  Shenzhen University, Shenzhen, China. Before joining Shenzhen
  University, he has been with the School of Information Science and
  Technology, Sun Yat-sen University, Guangzhou, China, since
  2000. His current research interests include multimedia forensics
  and security. He is an Associate Editor of the IEEE Transactions on
  Information Forensics and Security.
\end{IEEEbiography}


\vfill


\end{document}